\documentclass{article}

\usepackage{arxiv}

\usepackage[utf8]{inputenc} 
\usepackage[T1]{fontenc}    
\usepackage{hyperref}       
\usepackage{url}            
\usepackage{booktabs}       
\usepackage{amsfonts}       
\usepackage{nicefrac}       
\usepackage{microtype}      
\usepackage{lipsum}

\usepackage{amsmath, amssymb, amsfonts, amsthm, mathtools}
\usepackage{bm}               
\usepackage{physics}          

\usepackage{graphicx}
\usepackage{subcaption}
\usepackage{booktabs}
\usepackage{multirow}
\usepackage{array}
\usepackage{tabularx}

\usepackage[dvipsnames]{xcolor}
\usepackage{hyperref}
\hypersetup{
  colorlinks=true,
  linkcolor=MidnightBlue,
  citecolor=MidnightBlue,
  urlcolor=MidnightBlue
}
\usepackage[nameinlink,noabbrev]{cleveref}

\usepackage[numbers,sort&compress]{natbib}

\usepackage{algorithm}
\usepackage{algorithmic} 

\usepackage{listings}
\newcommand{\Ltbptt}{L_{\mathrm{tbptt}}}        
\newcommand{\hinel}{HINE\text{-}L2}            

\theoremstyle{plain}

\theoremstyle{definition}

\theoremstyle{remark}

\title{Stabilizing Autoregressive Forecasts in Chaotic Systems via Multi-rate Latent Recurrence}

\author{
  Mrigank Dhingra \\
  Department of Mechanical \& Aerospace Engineering\\
  University of Tennessee, Knoxville\\
  \texttt{mdhingra@vols.utk.edu} \\
  \And
  Omer San \\
  Department of Mechanical \& Aerospace Engineering\\
  University of Tennessee, Knoxville\\
  \texttt{osan@utk.edu}
}

\begin{document}
\maketitle

\begin{abstract}
Long-horizon autoregressive forecasting of chaotic dynamical systems remains challenging due to rapid error amplification and distribution shift: small one-step inaccuracies compound into physically inconsistent rollouts and collapse of large-scale statistics. We introduce \emph{MSR-HINE}, a hierarchical implicit forecaster that augments multiscale latent priors with \emph{multi-rate recurrent modules} operating at distinct temporal scales. At each step, coarse-to-fine recurrent states generate latent priors, an implicit one-step predictor refines the state with multiscale latent injections, and a gated fusion with posterior latents enforces scale-consistent updates; a lightweight hidden-state correction further aligns recurrent memories with fused latents. The resulting architecture maintains long-term context on slow manifolds while preserving fast-scale variability, mitigating error accumulation in chaotic rollouts.
Across two canonical benchmarks, MSR-HINE yields substantial gains over a U-Net autoregressive baseline: on Kuramoto--Sivashinsky it reduces end-horizon RMSE by \textbf{62.8\%} at $H{=}400$ and improves end-horizon ACC by \textbf{+0.983} (from $-0.155$ to $0.828$), extending the $\mathrm{ACC}\ge0.5$ predictability horizon from $241$ to $400$ steps; on Lorenz--96 it reduces RMSE by \textbf{27.0\%} at $H{=}100$ and improves end-horizon ACC by \textbf{+0.402} (from $0.144$ to $0.545$), extending the $\mathrm{ACC}\ge0.5$ horizon from $58$ to $100$ steps.
\end{abstract}
\section{Introduction}
\label{sec:intro}

Accurate long-horizon forecasting of chaotic dynamical systems is a central challenge in scientific computing, with applications spanning geophysical flows, reduced-order turbulence models, weather--climate prediction, and digital-twin surrogates for nonlinear partial differential equations (PDEs). In these settings, the goal is not merely to fit one-step transitions, but to generate stable rollouts that remain faithful to the system's evolving multiscale structure and long-time statistics. Autoregressive neural forecasters, which iteratively apply a learned one-step map, are appealing because they are simple to train, easy to deploy, and can be paired with expressive operator-learning backbones. However, when applied to chaotic systems, naïvely trained autoregressive models typically degrade rapidly: small one-step errors compound exponentially, driving rollouts off the attractor, corrupting low-frequency components, and ultimately producing unphysical trajectories.

A primary obstacle is \emph{distribution shift} between training and inference. During training, one-step predictors are commonly optimized under teacher forcing, where the model conditions on ground-truth states. At inference, the model conditions on its own predictions, so the input distribution drifts as rollout errors accumulate. This mismatch can cause error cascades even when the one-step mean-squared error is small. The resulting failure modes are well documented in chaotic dynamics: rollouts may become overly dissipative, lose phase coherence, collapse to spurious fixed points, or exhibit energy spectra that deviate markedly from the true dynamics. Improving rollout robustness therefore requires architectural and algorithmic mechanisms that explicitly account for multi-step error propagation while preserving physically meaningful structure.

Chaotic PDE-like systems also exhibit pronounced \emph{multiscale} and \emph{multi-timescale} behavior \cite{BIFERALE1999187, Wittenberg1999ScaleAS}. Large-scale (low-wavenumber) modes often evolve more slowly and encode the coarse organization of the state, while smaller scales fluctuate rapidly and are strongly coupled to the large scales through nonlinear interactions. A key practical implication is that forecasting errors are not uniform across scales: inaccuracies in coarse modes can induce persistent phase drift and global bias, whereas errors at fine scales can lead to excessive dissipation or spurious high-frequency noise. This suggests that models should represent and propagate information across a hierarchy of scales, and that slower components should carry memory over longer horizons.

\begin{figure}
    \centering
    \includegraphics[width=\linewidth]{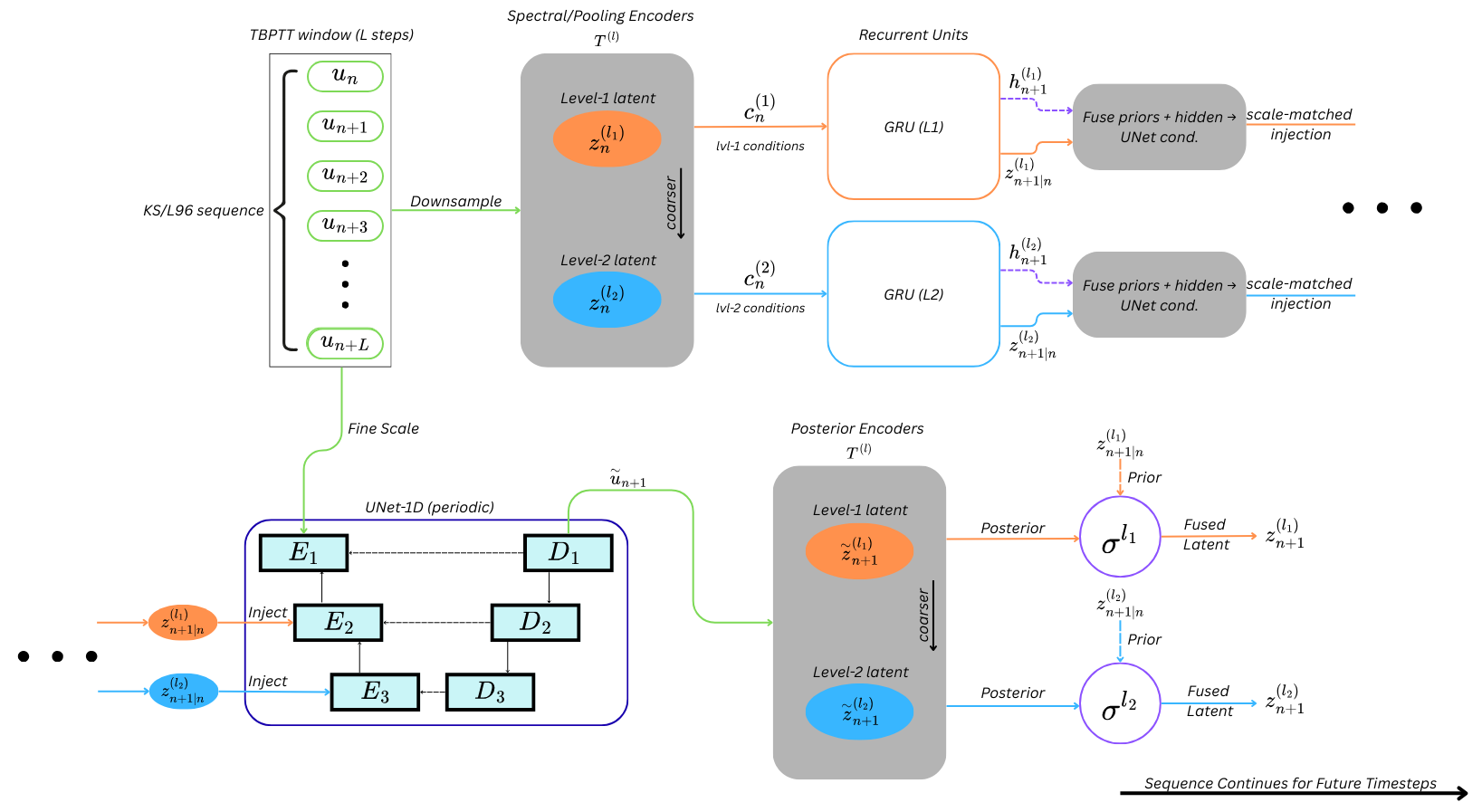}
    \caption{\textbf{MSR-HINE architecture for one-step forecasting within TBPTT.}
A length-$\Ltbptt$ window from the KS/L96 trajectory is encoded into a hierarchy of multiscale latents $\{z_n^{(l)}\}_{l=1}^L$ using spectral/pooling encoders $T^{(l)}$ (coarser with increasing level). At each level, a recurrent unit (GRU) propagates a multi-rate hidden state $h_{n+1}^{(l)}$ and produces a latent prior $z_{n+1|n}^{(l)}$, which is fused with the hidden state to form scale-matched conditioning inputs for a periodic 1D U-Net. The U-Net predicts the next state $\hat{u}_{n+1}$ while injecting the multiscale priors at corresponding encoder depths. Posterior latents $\{\tilde{z}_{n+1}^{(l)}\}$ are then extracted from $\hat{u}_{n+1}$ via the same hierarchy of encoders and combined with the priors through a gated fusion module $\sigma^{(l)}$ to obtain fused latents $\{z_{n+1}^{(l)}\}$ used for the next step. The procedure is iterated autoregressively to generate long-horizon rollouts.}
    \label{fig:msr_hine_schematic}
\end{figure}

Motivated by these observations, we develop a multiscale autoregressive forecaster that combines (i) an \emph{implicit} one-step predictor that refines the next state conditioned on latent priors, and (ii) a \emph{multi-rate recurrent hierarchy} that propagates latent dynamics at multiple temporal scales (see Figure ~\ref{fig:msr_hine_schematic}). Our approach builds on the view that a stable chaotic forecast should maintain a coherent representation of slow manifolds while allowing fast fluctuations to evolve conditioned on the slow context. To this end, we construct a hierarchy of latent variables capturing progressively coarser content of the state, and equip each latent level with its own recurrent module. Coarser recurrent modules update at slower rates and carry long-term memory, while finer modules update more frequently to capture rapid dynamics. At each time step, recurrent modules produce \emph{latent priors} for the next state; an implicit predictor (a one-step U-Net-style model with multiscale injections) then produces the next state estimate. We additionally extract \emph{posterior} latents from the newly obtained next state and perform a gated fusion between prior and posterior latents to enforce scale-consistent updates. A lightweight hidden-state correction aligns recurrent memories with the fused latents, further stabilizing long rollouts.

We refer to the resulting framework as \emph{MSR-HINE} (Multi-Scale Recurrent Hierarchical Implicit Neural forecaster). The method is designed to be practical: it adds only modest recurrent capacity per latent level, preserves a one-step prediction interface, and can be trained using standard truncated backpropagation through time (TBPTT) with lightweight regularizers and spectral/energy-aware losses. Crucially, the hierarchical recurrence enables the model to maintain long-term context at coarse scales and to ``nudge'' finer predictions when they begin to drift, functioning analogously to a predictor--corrector mechanism distributed across scales.

We evaluate MSR-HINE on two canonical chaotic benchmarks: the one-dimensional Kuramoto--Sivashinsky (KS) system and the Lorenz--96 (L96) ring model. These benchmarks expose complementary difficulties: KS exhibits rich spatiotemporal chaos with a clear spectral interpretation, while L96 presents strong nonlinearity and rapid error growth in a compact state space. Across both systems, we assess not only pointwise forecast error but also physically meaningful diagnostics including low-wavenumber spectral accuracy, band-energy drift, and long-time statistical consistency. Our experiments show that multi-rate latent recurrence and scale-consistent fusion substantially improve forecast stability and long-horizon accuracy relative to strong autoregressive baselines.

In summary, our main contributions are:
\begin{itemize}
    \item \textbf{Multi-rate latent recurrence for chaotic forecasting.} We introduce a hierarchy of recurrent modules operating on coarse-to-fine latents to capture multi-timescale dependencies and improve long-horizon stability.
    \item \textbf{Scale-consistent implicit refinement.} We combine latent priors with an implicit one-step predictor and a gated prior--posterior fusion mechanism, complemented by a hidden-state correction that stabilizes the recurrent memories.
    \item \textbf{Comprehensive evaluation on KS and L96.} We demonstrate improved long-horizon rollouts and better spectral/energy and statistical fidelity, and provide ablations isolating the roles of recurrence, fusion, conditioning, and correction.
\end{itemize}

The remainder of the paper is organized as follows. \Cref{sec:related} reviews related work in neural operator forecasting, autoregressive modeling, and multi-timescale recurrence. \Cref{sec:setup} defines the forecasting tasks and datasets. \Cref{sec:method} presents MSR-HINE, including the latent hierarchy, conditioning, multi-rate recurrence, and implicit refinement with fusion and correction. \Cref{sec:exps} describes experimental details and baselines, and \Cref{sec:results} reports quantitative and qualitative results with ablations and diagnostics. Finally, \Cref{sec:discussion} discusses limitations and future directions, and \Cref{sec:conclusion} concludes.

\section{Related Work}
\label{sec:related}

Learning accurate surrogates for PDEs has a long history spanning classical reduced-order modeling and modern deep learning. Early physics-informed approaches (PINNs) optimize neural representations of solutions by penalizing PDE residuals and constraints, enabling flexible inverse-problem and sparse-data settings \cite{RAISSI2019686}. Variational and weak-form formulations, such as the Deep Ritz method and Deep Galerkin-type approaches, further connect neural PDE solvers to energy minimization and Galerkin principles \cite{11a35c8971ea4ca6af3db3c5545af9ed,repec:arx:papers:1708.07469}. While these methods can be powerful, long-horizon forecasting in chaotic regimes often benefits from operator-learning viewpoints that learn \emph{families} of solution maps and generalize across input functions and parameters.

Operator learning frameworks aim to approximate mappings between function spaces (e.g., initial conditions, coefficients, or forcing fields to solutions). DeepONet introduced a branch--trunk architecture to learn nonlinear operators from paired function observations and has become a widely used baseline for parametric PDE learning \cite{Lu_2021}. A complementary formulation is provided by neural operator theory, which motivates discretization-invariant architectures that learn operators directly and generalize across meshes \cite{10.5555/3648699.3648788}. This perspective has catalyzed a large family of practical neural operators for PDE surrogates.

Among the most widely adopted architectures is the Fourier Neural Operator (FNO), which parameterizes global convolutions in spectral space via truncated Fourier modes, offering strong performance on canonical PDE benchmarks and efficient training/inference \cite{li2021fourierneuraloperatorparametric}. Subsequent work extends FNO-style operators in multiple directions: multiresolution U-shaped neural operators improve representation of multiscale structure and enable coarse-to-fine information flow \cite{rahman2023uno}; wavelet neural operators introduce locality and multiscale adaptivity via wavelet transforms \cite{Tripura2022WaveletNO}; and convolutional neural operator variants emphasize localized kernels and improved resolution generalization in challenging regimes \cite{fan2025convolutionalneuraloperatorbasedtransferlearningsolving}. Other extensions target nonuniform sampling and irregular discretizations, addressing practical settings where data are observed on nonuniform grids or varying meshes \cite{liu2023nunogeneralframeworklearning}.

Transformers have also been adapted for operator learning to better model nonlocal interactions and long-range couplings. Operator transformer families (e.g., Galerkin- or Fourier-inspired attention) connect attention mechanisms to projection/weak-form ideas and demonstrate strong results across PDE families \cite{Cao2021transformer}. More recent scalable transformer-based neural operators (e.g., graph/geometry-aware operator transformers) expand these ideas to broader domains and improved generalization \cite{liu2025geometryinformedneuraloperatortransformer}. In parallel, physics-informed neural operators incorporate PDE structure (e.g., residual constraints) into supervised operator learning, aiming to combine the data-efficiency and physical bias of PINNs with the generalization and scalability of neural operators \cite{li2023physicsinformedneuraloperatorlearning}.

Beyond grid-based PDE surrogates, graph-based physics simulators provide an important complementary direction. MeshGraphNets demonstrate that message-passing architectures can learn accurate surrogates on unstructured meshes, highlighting the role of inductive biases for discretization generalization \cite{pfaff2021learningmeshbasedsimulationgraph}. Equivariant and geometry-aware operator learning further leverages symmetry and geometric structure to improve sample efficiency and robustness across domains \cite{catalani2025geometryawareinferencesteady,li2023geometryinformedneuraloperatorlargescale}. These directions are increasingly relevant as surrogate models move from idealized periodic grids to complex geometries and heterogeneous discretizations.

Finally, standardized datasets and benchmarks are central for reproducible evaluation of PDE surrogates, especially when comparing generalization across parameters, resolutions, and rollout horizons. PDEBench provides a broad benchmark suite with datasets, baselines, and protocols for operator-learning models and related PDE surrogates \cite{takamoto2024pdebenchextensivebenchmarkscientific}. Additional benchmark efforts and curated datasets further support systematic comparisons and ablation-driven analysis of architectures and training choices \cite{gupta2022towards,serrano2023operatorlearningneuralfields}. Taken together, this literature provides both the architectural building blocks and the evaluation methodology needed to assess stability and long-horizon fidelity of learned surrogates for chaotic dynamical systems \cite{DBLP:journals/corr/abs-2108-08481,li2021fourierneuraloperatorparametric,Lu_2021}.

A common strategy for forecasting chaotic dynamical systems is \emph{autoregressive} (AR) rollouts: a learned one-step map is iterated forward by feeding predictions back as inputs. In practice, AR training is often performed under teacher forcing with truncated backpropagation through time (TBPTT), and the learned forecaster is then deployed in closed loop, where distribution shift can rapidly amplify small one-step errors. This \emph{exposure bias} has been studied extensively in sequence modeling, motivating curriculum-style training schemes and adversarial matching of free-running dynamics \cite{bengio2015scheduledsamplingsequenceprediction, lamb2016professorforcingnewalgorithm, pmlr-v28-sutskever13, 6795228}.

For chaotic systems, the literature documents that minimizing one-step error alone is frequently insufficient for stable long-horizon behavior, even when short-term forecasts look accurate \cite{Vlachas2018LSTMChaos, VLACHAS2020191, Lusch_2018}. Classical and modern approaches therefore emphasize \emph{closed-loop stability} and \emph{multi-step consistency}. Reservoir computing (RC) and echo-state methods have shown strong long-horizon skill on canonical chaotic benchmarks, including Lorenz--96 and Kuramoto--Sivashinsky-type dynamics, by leveraging high-dimensional recurrent state spaces \cite{Pathak_2017, PhysRevLett.120.024102, PhysRevE.98.012215}. Hybrid strategies that blend mechanistic structure with learned corrections further improve robustness when partial physics is known or when model error dominates \cite{Pathak_2018, rudy2016datadrivendiscoverypartialdifferential, PhysRevFluids.6.034402}. In geophysical and multiscale settings, hierarchical and multi-resolution designs (e.g., multi-model deep learning stacks or scale-separated training) are frequently used to address disparate time scales and unresolved variability \cite{Chattopadhyay_2020, https://doi.org/10.1029/2021MS002712, https://doi.org/10.1029/2021MS002572}.

A complementary line of work targets \emph{statistical} and \emph{invariant-measure} fidelity, recognizing that for ergodic chaotic systems, matching long-run statistics and attractor geometry may matter more than pointwise trajectory tracking at long horizons. Recent methods regularize training by comparing forecast-generated distributions to data distributions, improving stability and long-term climatology in closed loop \cite{schiff2024dyslimdynamicsstablelearning, jiang2024trainingneuraloperatorspreserve}. Relatedly, latent-state and operator-learning frameworks can stabilize AR rollouts by constraining the forecast to evolve on structured latent manifolds (e.g., Koopman-inspired coordinates) and by using hierarchical priors/posteriors to inject multiscale context \cite{Lusch_2018, jiang2025hierarchicalimplicitneuralemulators, brandstetter2023messagepassingneuralpde, DBLP:journals/corr/abs-2108-08481}. In this paper, we build on these insights by explicitly allocating recurrent memory across latent scales to reduce drift and improve temporal coherence under AR deployment, while retaining the efficiency of one-step training and the practicality of TBPTT.

A long line of work frames forecasting as \emph{latent-state} inference in a dynamical system, where an unobserved state $z_t$ evolves with (possibly stochastic) dynamics and generates observations $u_t$. Modern deep latent-variable sequence models combine amortized variational inference with recurrent or state-space structure to learn both (i) a transition model in latent space and (ii) a decoder back to observation space. Early and influential examples include the Variational Recurrent Neural Network (VRNN) \cite{chung2016recurrentlatentvariablemodel} and deep nonlinear state-space formulations such as the Deep Kalman Filter (DKF) \cite{krishnan2015deepkalmanfilters} and Deep Markov Models (DMM) \cite{krishnan2016structuredinferencenetworksnonlinear}, which explicitly model latent transitions and observation likelihoods. Deep Variational Bayes Filters (DVBF) \cite{karl2017deepvariationalbayesfilters} further emphasize filtering-style inference with learned dynamics and recognition models.

For continuous-time and stiff/chaotic dynamics, latent dynamics are often parameterized by differential equations, enabling time-adaptive integration and improved long-horizon structure. Latent ODE models \cite{rubanova2019latentodesirregularlysampledtime} represent $z(t)$ with neural ODE dynamics and infer initial conditions from partial observations; ODE2VAE \cite{yıldız2019ode2vaedeepgenerativesecond} extends this idea to second-order latent dynamics suitable for oscillatory systems. Stochasticity and model error can be represented via latent SDEs \cite{zeng2024latentsdeshomogeneousspaces}, which are especially relevant for chaotic regimes where unresolved scales induce effective noise. In neuroscience and low-dimensional manifold modeling, LFADS \cite{Pandarinath2018LFADS} demonstrates how structured latent dynamical priors can yield stable multi-step reconstructions and forecasts from noisy measurements.

A complementary direction focuses on \emph{hierarchical} or \emph{multi-timescale} recurrence: instead of a single recurrent module, the model maintains multiple recurrent states operating at different update rates or temporal receptive fields. Clockwork RNNs \cite{neil2016phasedlstmacceleratingrecurrent} partition hidden units into bands updated at different periods, while Hierarchical Multiscale RNNs (HM-RNN) \cite{chung2017hierarchicalmultiscalerecurrentneural} learn boundary variables to adaptively allocate computation across temporal scales. These multi-rate ideas align closely with chaotic PDE forecasting, where slow manifold components and fast small-scale components coexist; our multi-rate latent recurrence can be viewed as a structured hybrid of latent-state modeling and explicit timescale separation.

Predict--correct viewpoints are central to sequential estimation: one alternates between (i) a \emph{forecast} (predict) step that propagates uncertainty through a dynamical model and (ii) an \emph{analysis} (correct) step that incorporates observations. The Kalman filter \cite{kalman_filter} is the classical linear-Gaussian instance of this paradigm, and many nonlinear/chaotic extensions preserve the same two-stage structure. In large-scale geophysical systems, ensemble methods approximate uncertainty with Monte Carlo particles, leading to the Ensemble Kalman Filter (EnKF) \cite{1994JGR....9910143E} and practical analysis schemes \cite{AnalysisSchemeintheEnsembleKalmanFilter}; localized/square-root variants such as the Ensemble Square Root Filter \cite{AGSIBasedCoupledEnSRFEn3DVarHybridDataAssimilationSystemfortheOperationalRapidRefreshModelTestsataReducedResolution} and LETKF \cite{hunt2006efficientdataassimilationspatiotemporal} improve stability and scalability by controlling sampling error.

Beyond EnKF-style Gaussian updates, particle filtering methods (e.g., bootstrap/SIR) \cite{Gordon1993ParticleFilter} handle strongly nonlinear/non-Gaussian regimes via importance sampling and resampling, while sigma-point approaches like the Unscented Kalman Filter (UKF) \cite{882463} provide deterministic moment propagation. Variational data assimilation (e.g., incremental 4D-Var) \cite{https://doi.org/10.1002/qj.49712051912} can also be interpreted as a predict--correct procedure over a time window by optimizing a trajectory consistent with both model and observations. More recently, hybrid approaches learn parts of the forecast model or the correction operator using neural networks while retaining the predict--correct scaffold; for example, DA+ML hybrids on Lorenz-style systems highlight how learned corrections can improve long-range skill \cite{BRAJARD2020101171}. Fully learned filtering operators that map sequences of observations to state estimates (or Kalman gains) further connect to this view, e.g., KalmanNet \cite{Revach_2022}.

In our setting, the “correct” signal is implemented in latent space (via fusion/coupling mechanisms) to stabilize autoregressive rollouts: multi-rate latent recurrence supplies a structured forecast prior, while posterior/fusion acts as a corrective mechanism that enforces multiscale consistency. This framing makes it natural to compare against (and borrow tools from) filtering/assimilation literature when designing losses, schedules (teacher forcing vs. free-rollout), and stability controls for chaotic forecasting.

\section{Problem Setup}
\label{sec:setup}
\subsection{Kuramoto--Sivashinsky (KS)}
\label{sec:ks_setup}

We consider the one-dimensional KS dynamics on a periodic domain
$x \in [0,2\pi L)$ with $L=16$, discretized on a uniform grid of $N=128$ points with spacing
$\Delta x = 2\pi L/N$. We evolve the state $u(x,t)$ using a pseudo-spectral method based on
the real FFT (rFFT). Let $\hat{u}(k,t)$ denote the rFFT coefficients and let
$k$ be the corresponding angular wavenumbers computed from rFFT frequency bins.

\paragraph{Spectral form and de-aliasing.}
The nonlinear term is computed in Fourier space as
\begin{equation}
\widehat{\mathcal{N}(u)}(k) \;=\; -\tfrac{1}{2} i k \, \widehat{u^2}(k),
\end{equation}
where $\widehat{u^2}$ is obtained by transforming to physical space, forming $u^2$,
and transforming back. To control aliasing errors from the quadratic nonlinearity,
we use the standard $3/2$ zero-padding strategy: the spectrum is padded to a
physical grid of length $M = \lfloor 3N/2 \rfloor$, the product $u^2$ is formed on that grid,
and the resulting spectrum is truncated back to the first $N/2+1$ rFFT modes.
(We also support a cheaper $2/3$-rule masking variant, though the default is $3/2$ padding.)

\paragraph{Time integration (IMEX--BDF2).}
Time integration is performed with an IMEX BDF2 scheme: the linear part is handled implicitly
in Fourier space, while the nonlinear term is treated explicitly. The linear operator is applied
diagonally in Fourier space with symbol
\begin{equation}
\mathcal{L}(k) \;=\; -(k^2) + (k^4).
\end{equation}
We bootstrap the first step using a backward-Euler predictor followed by a Crank--Nicolson
linear update with a midpoint nonlinearity. For subsequent steps, the BDF2/AB2-style update in
Fourier space takes the form
\begin{equation}
\hat{u}^{n+1}
=
\frac{
4\hat{u}^{n} - \hat{u}^{n-1} + 2\Delta t \left( 2\hat{\mathcal{N}}^{n} - \hat{\mathcal{N}}^{n-1} \right)
}{
3 + 2\Delta t \, \mathcal{L}(k)
},
\end{equation}
where $\hat{\mathcal{N}}^{n}$ denotes the de-aliased nonlinear term computed from $\hat{u}^{n}$.
To remove mean drift, we enforce a zero-mean constraint by setting the $k=0$ Fourier mode to zero
at every step.

\paragraph{Initial conditions.}
Each trajectory is initialized from a randomized smooth field formed by a small set of low modes:
\begin{equation}
u_0(x) = \sum_{j \in \{1,2,3,4\}} a_j \cos\!\left(\frac{j x}{L}\right)\left(1 + \sin\!\left(\frac{j x}{L}\right)\right),
\qquad a_j \sim \mathcal{U}(0,1),
\end{equation}
and then rescaled so that $\max_x |u_0(x)| = 2.0$.

\paragraph{Dataset construction and burn-in.}
We generate $S$ independent rollouts by integrating each initial condition with a fixed step size
$\Delta t = 0.10$ up to a terminal time $T_f$ (e.g., $T_f=200$ in our default configuration).
To reduce transient dependence on initialization and to better sample the attractor,
we discard an initial burn-in interval of length $T_{\mathrm{burn}} = 50.0$ time units.
The final dataset stores only the post burn-in states:
\begin{equation}
\mathbf{U} \in \mathbb{R}^{S \times T_b \times N},
\qquad
T_b = \left\lfloor \frac{T_f}{\Delta t}\right\rfloor + 1 - \left\lfloor \frac{T_{\mathrm{burn}}}{\Delta t}\right\rfloor,
\end{equation}
along with the post burn-in time vector and the state at the burn-in time (used as a canonical
starting state for forecasting experiments). All arrays are saved in a single compressed \texttt{.npz}
file together with metadata (grid/domain parameters, integrator name, de-aliasing choice, and burn-in).

\subsection{Lorenz--96 (L96)}
\label{sec:setup:l96}

We consider the standard Lorenz--96 system on a periodic ring of $N$ state variables,
\begin{equation}
\frac{d x_i}{dt} = (x_{i+1} - x_{i-2})\,x_{i-1} - x_i + F,\qquad i=1,\dots,N,
\end{equation}
with cyclic indexing $x_{i\pm k}\equiv x_{(i\pm k \;\mathrm{mod}\; N)}$. In our implementation, the right-hand side is realized using periodic shifts (rolls) to form $(x_{i+1},x_{i-1},x_{i-2})$ and the additive forcing term $F$.

\paragraph{Parameters and time window.}
The generator is configured with state dimension $N=40$ and forcing $F=8.0$.
We generate $N_{\mathrm{traj}}=100$ independent trajectories and store $N_{\mathrm{saved}}=1000$ uniformly spaced snapshots over the \emph{visible} time interval $t\in[T_{\mathrm{vis,start}},T_{\mathrm{vis,end}}]=[0,50]$. 
Integration uses a fixed RK4 time step $\Delta t_{\mathrm{int}}=0.01$. 
A short spin-up (burn-in) interval is included by integrating from
\begin{equation}
t_{\mathrm{int,start}} = T_{\mathrm{vis,start}} - \alpha_{\mathrm{pre}}(T_{\mathrm{vis,end}}-T_{\mathrm{vis,start}})
\end{equation}
with $\alpha_{\mathrm{pre}}=0.05$, i.e., $t_{\mathrm{int,start}}=-2.5$ for the above window, and ending at $t_{\mathrm{int,end}}=T_{\mathrm{vis,end}}$. 
The total number of RK4 steps is computed as $\lceil (t_{\mathrm{int,end}}-t_{\mathrm{int,start}})/\Delta t_{\mathrm{int}}\rceil$.

\paragraph{Numerical integration.}
For each trajectory, we integrate the L96 dynamics using a classical fourth-order Runge--Kutta (RK4) method. The RK4 update is implemented explicitly via $(k_1,k_2,k_3,k_4)$ evaluations of the L96 right-hand side, and the full trajectory is generated efficiently with a loop. 

\paragraph{Initial conditions.}
Initial conditions are drawn near the equilibrium $x_i=F$ with small i.i.d.\ Gaussian perturbations of standard deviation $10^{-2}$:
\begin{equation}
x_i(0) = F + 0.01\,\xi_i,\qquad \xi_i\sim \mathcal{N}(0,1).
\end{equation}

\paragraph{Uniform resampling to a fixed-length sequence.}
To obtain a consistent sequence length across trajectories, we resample the integrated (spin-up + visible) solution onto $N_{\mathrm{saved}}$ linearly spaced target times in $[T_{\mathrm{vis,start}},T_{\mathrm{vis,end}}]$. This is done by masking the integrated time grid to the visible interval and applying 1D interpolation independently to each component $x_i(t)$.

\subsection{Forecasting task}
\label{sec:forecast_task}

Given a trajectory of system states $\{u_t\}_{t\ge 0}$ sampled at a fixed time increment $\Delta t$,
our objective is \emph{long-horizon autoregressive forecasting} for chaotic dynamics.
We learn a one-step predictor $f_\theta$ that advances the state by one sampling interval,
\begin{equation}
\hat{u}_{t+1} = f_\theta(u_t),
\end{equation}
where $u_t\in\mathbb{R}^N$ denotes the full system state (KS field on an $N$-point grid or L96 ring state).
At inference time, multi-step forecasts are obtained by iterating the learned map in closed loop:
\begin{equation}
\hat{u}_{t+h} = f_\theta(\hat{u}_{t+h-1}), \qquad h = 1,2,\dots,H,
\end{equation}
with $\hat{u}_t = u_t$ at the rollout start time.
Because the systems are chaotic, stable long-horizon performance requires controlling error accumulation
under repeated self-feeding. Accordingly, we evaluate models using both one-step accuracy and the
growth of forecast error over free-rollouts of length $H$ (see Sec.~\ref{sec:eval_metrics} for protocols and metrics).

\section{Method: MSR-HINE}
\label{sec:method}

\subsection{Overview}
\label{sec:method_overview}

We learn an autoregressive one-step forecaster for a discrete-time system with state $u_t\in\mathbb{R}^N$.
MSR-HINE couples three components (expanded in Secs.~\ref{sec:latent_encoders}--\ref{sec:predict_correct}):

(i) \textbf{Latent hierarchy.} A multi-level encoder $\mathcal{T}^{(\ell)}$ produces scale-ordered latents
$z_t^{(\ell)}=\mathcal{T}^{(\ell)}(u_t)$, $\ell=1,\dots,L$.

(ii) \textbf{Multi-rate recurrent priors.} Per-level GRUs update hidden states $h_{t+1}^{(\ell)}$ at stride $s_\ell$
and read out latent priors $z_{t+1|t}^{(\ell)}$; slow levels evolve smoothly, fast levels react quickly.

(iii) \textbf{Predict--correct in latent space.} A U-Net predicts $\hat{u}_{t+1}$ conditioned on
$\{z_{t+1|t}^{(\ell)},h_{t+1}^{(\ell)}\}_{\ell=1}^L$. Posterior latents from $u_{t+1}^{\mathrm{post}}
\in\{u_{t+1},\hat{u}_{t+1}\}$ are fused with the priors by a learned gate to yield $z_{t+1}^{(\ell)}$;
a light hidden correction aligns $h_{t+1}^{(\ell)}$ to the fused latents.

\subsection{Spectral latent encoders for KS and L96}
\label{sec:latent_encoders}

Both KS and L96 are periodic 1D systems (field on a ring or variables on a ring),
which makes Fourier representations a natural, physically meaningful latent space.
We construct hierarchical latents by retaining only low wavenumbers at coarser levels.

\paragraph{Low-wavenumber Fourier slices.}
Let $u_t \in \mathbb{R}^N$ be the state at time $t$ and $U_t=\mathrm{rFFT}(u_t)$ its discrete real FFT coefficients.
For each level $\ell$ we choose a cutoff $K_\ell$ and define a real-valued latent by packing real and imaginary parts:
\begin{equation}
z_t^{(\ell)} \;=\;
\Big[\;\Re(U_t[0{:}K_\ell]);\; \Im(U_t[0{:}K_\ell])\;\Big]\;\in\;\mathbb{R}^{2K_\ell}.
\end{equation}
This yields a nested hierarchy: coarser levels retain fewer modes and thus evolve on slower timescales.

\paragraph{KS derivative-norm features (auxiliary).}
For KS, we additionally exploit the availability of spatial derivatives to form compact diagnostics of
small-scale activity. Using spectral multipliers, the $m$-th derivative is computed as
\begin{equation}
\partial_x^m u_t \;=\; \mathrm{irFFT}\!\big((\mathrm{i}k)^m \odot U_t\big),
\end{equation}
and we use L2 norms $\|\partial_x^m u_t\|_2$ (for selected orders $m$) as scalar summaries.
These quantities are used as part of the conditioning features (Sec.~\ref{sec:conditioning_features})
rather than as latents to be propagated.

\subsection{Conditioning features}
\label{sec:conditioning_features}

The recurrent priors $\mathcal{R}^{(\ell)}$ do not observe $u_t$ directly; instead they receive compact
conditioning vectors $c_t^{(\ell)}$ designed to (i) inform each latent level about the present state and
(ii) emphasize changes over time via finite differences.
We maintain a small cache of previous-step statistics to construct temporal increments
$\Delta(\cdot)_t = (\cdot)_t - (\cdot)_{t-1}$.

\paragraph{KS conditioners (field + derivative structure).}
For KS we use two complementary feature groups:
\begin{itemize}
\item \textbf{Mid-level conditioning} from coarse spatial averages:
partition the 1D domain into bins and compute pooled averages $p_t$ (avg-pooling of $u_t$),
then form $[p_t,\, \Delta p_t]$ and project to $c_t^{(1)}$. We also inject a top-down summary of the coarse latent
via a learned linear map of $z_t^{(2)}$.
\item \textbf{Coarse-level conditioning} from spectral and dissipative summaries:
use a low-$k$ Fourier slice of $u_t$ together with derivative norms
$\|\partial_x^2 u_t\|_2$, $\|\partial_x^4 u_t\|_2$ and their increments.
The resulting concatenated vector is projected to $c_t^{(2)}$.
\end{itemize}
These features jointly encode large-scale content (pooled bins and low-$k$ modes) and the level of
small-scale activity (derivative norms), which is crucial for stabilizing rollouts in KS.

\paragraph{L96 conditioners (ring statistics + spectral-energy content).}
For L96 we use analogous ring-consistent features:
\begin{itemize}
\item \textbf{Mid-level conditioning} from pooled groups along the ring:
compute group means $p_t$ by averaging contiguous subsets of components of $u_t$,
concatenate $[p_t,\, z_t^{(2)},\, \Delta p_t]$, and project to $c_t^{(1)}$.
\item \textbf{Coarse-level conditioning} from global and spectral-energy summaries:
include a scaled low-$k$ Fourier slice, global mean and standard deviation, their increments,
and multi-band spectral energies $E_{b,t}=\sum_{k\in\mathcal{B}_b} |U_t(k)|^2$ (excluding DC),
together with $\Delta E_{b,t}$ and the total spectral energy and its increment.
\end{itemize}

\subsection{Multi-rate latent recurrence}
\label{sec:msr}

A central goal of MSR-HINE is to endow the latent hierarchy with \emph{memory} on multiple timescales while keeping the update cost and optimization stable. We achieve this via a multi-rate recurrent prior in latent space: each latent level $\ell \in \{1,\dots,L\}$ has its own recurrent state $h_t^{(\ell)}$ and its own update stride $s_\ell \in \mathbb{N}$ (coarser levels typically use larger $s_\ell$).

\paragraph{Per-level recurrent prior.}
Given a conditioning vector $c_t^{(\ell)} \in \mathbb{R}^{d_\text{in}^{(\ell)}}$ (Sec.~\ref{sec:conditioning_features}), the level-$\ell$ module produces a prior latent for the \emph{next} time index together with an updated hidden state:
\begin{equation}
\big(z_{t+1|t}^{(\ell)},\, h_{t+1}^{(\ell)}\big)
=
\mathcal{R}^{(\ell)}\!\big(h_t^{(\ell)},\, c_t^{(\ell)}\big),
\label{eq:per_level_prior}
\end{equation}
where $\mathcal{R}^{(\ell)}$ is implemented as a GRU with a linear readout to the latent dimension.

\paragraph{Multi-rate scheduling.}
At level $\ell$, a \emph{full} recurrent update is performed only on aligned steps,
\begin{equation}
t \in \mathcal{A}_\ell
\;\;\Longleftrightarrow\;\;
t \equiv 0 \!\!\!\pmod{s_\ell},
\label{eq:aligned_steps}
\end{equation}
with an explicit alignment at $t=0$ to ensure that each chunk starts from a consistent prior update. On non-aligned steps, we avoid unnecessary fast oscillations in coarse memories by using either (i) a strict hold (no hidden update) or (ii) a lightweight exponential moving-average (EMA) micro-update (Skip-RNN style). Concretely, let $\tilde{h}_{t+1}^{(\ell)}$ be the hidden state that would be obtained by a standard GRU transition on input $c_t^{(\ell)}$. Then on off-stride steps we set
\begin{equation}
h_{t+1}^{(\ell)} \;=\; (1-\alpha_\ell)\,h_t^{(\ell)} \;+\; \alpha_\ell\,\tilde{h}_{t+1}^{(\ell)},
\qquad \alpha_\ell \in (0,1],
\label{eq:ema_micro_update}
\end{equation}
with $\alpha_\ell$ chosen to be small for slow levels (e.g., proportional to $1/s_\ell$). The corresponding prior latent $z_{t+1|t}^{(\ell)}$ is always obtained from the (possibly held or EMA-updated) hidden state via a readout map:
\begin{equation}
z_{t+1|t}^{(\ell)} \;=\; \rho^{(\ell)}\!\big(h_{t+1}^{(\ell)}\big).
\label{eq:latent_readout}
\end{equation}
This design enforces a separation of timescales: fine levels can react quickly while coarse levels evolve smoothly and provide stabilizing context.
Multi-rate recurrence plays a similar role to multi-timescale integrators and hierarchical filters: coarse memories track slow ``macro'' evolution and provide long-horizon coherence, while fine memories capture fast fluctuations. In contrast to purely autoregressive state-space updates, this structure allows the predictor to condition on both current information and persistent latent context without requiring prohibitively long backpropagation through time.

\subsection{Predict--correct fusion and hidden coupling}
\label{sec:predict_correct}

MSR-HINE adopts a predict--correct perspective in latent space: we first form a \emph{prior} latent prediction using the multi-rate recurrent dynamics (predict), then form a \emph{posterior} latent by encoding the next state (correct), and finally fuse the two using a learned gate.

\paragraph{Prior prediction.}
At time $t$, the multi-rate recurrence produces the next-step latent priors $\{z_{t+1|t}^{(\ell)}\}_{\ell=1}^L$ and updated hidden states $\{h_{t+1}^{(\ell)}\}_{\ell=1}^L$ according to Eq.~\eqref{eq:per_level_prior}--\eqref{eq:latent_readout}.

\paragraph{State prediction with latent/hidden injections.}
The one-step state predictor (e.g., a 1D U-Net) advances the state using the current state and the recurrent priors. In our implementation, we inject both the latent priors and the corresponding hidden states into the predictor:
\begin{equation}
\hat{u}_{t+1}
=
\mathcal{G}_\theta\!\Big(
u_t,\,
\{\,[z_{t+1|t}^{(\ell)};\, \tilde{h}_{t+1}^{(\ell)}]\,\}_{\ell=1}^L
\Big),
\label{eq:state_predictor_injection}
\end{equation}
where $[\,\cdot;\cdot\,]$ denotes concatenation and $\tilde{h}_{t+1}^{(\ell)}$ denotes a normalized hidden feature (e.g., LayerNorm applied to $h_{t+1}^{(\ell)}$). The predictor is typically parameterized as a residual update $\hat{u}_{t+1}=u_t+\Delta u_\theta(\cdot)$ to encourage stable one-step corrections.

\paragraph{Posterior construction.}
To correct the latent trajectory, we compute posterior latents at the \emph{same time index} $t{+}1$ by applying the latent encoder to a field $u_{t+1}^{\mathrm{post}}$:
\begin{equation}
z_{t+1}^{\mathrm{post},(\ell)} = \mathcal{T}^{(\ell)}(u_{t+1}^{\mathrm{post}}),
\qquad
u_{t+1}^{\mathrm{post}} \in \{u_{t+1},\, \hat{u}_{t+1}\}.
\label{eq:posterior_latents}
\end{equation}
During teacher-forced training, $u_{t+1}^{\mathrm{post}}=u_{t+1}$ (ground truth); during free-rollout (or free-tail segments), $u_{t+1}^{\mathrm{post}}=\hat{u}_{t+1}$.

\paragraph{Gated fusion.}
We fuse the prior and posterior latents elementwise using a learned gate:
\begin{align}
g_{t+1}^{(\ell)} &=
\sigma\!\Big(
W^{(\ell)}[z_{t+1|t}^{(\ell)};\, z_{t+1}^{\mathrm{post},(\ell)}] + b^{(\ell)}
\Big),
\label{eq:gate}\\
z_{t+1}^{(\ell)} &=
g_{t+1}^{(\ell)} \odot z_{t+1}^{\mathrm{post},(\ell)}
+ \big(1-g_{t+1}^{(\ell)}\big) \odot z_{t+1|t}^{(\ell)}.
\label{eq:fusion}
\end{align}
This operation can be interpreted as a learned, diagonal gain that interpolates between ``trust the dynamics'' (prior) and ``trust the encoded evidence'' (posterior), analogous to a simplified filtering update.

\paragraph{Hidden coupling via latent innovation.}
A key practical issue is that the recurrent hidden state $h_{t+1}^{(\ell)}$ was produced to support the prior $z_{t+1|t}^{(\ell)}$, but after fusion we wish the internal memory to remain consistent with the corrected latent $z_{t+1}^{(\ell)}$. We therefore apply a corrective coupling based on the latent innovation
\begin{equation}
\delta z_{t+1}^{(\ell)} = z_{t+1}^{(\ell)} - z_{t+1|t}^{(\ell)}.
\label{eq:innovation}
\end{equation}
Specifically, we map $\delta z_{t+1}^{(\ell)}$ to the hidden dimension and add it to the hidden state:
\begin{equation}
h_{t+1}^{(\ell)}
\leftarrow
h_{t+1}^{(\ell)} + \alpha_{\mathrm{corr}}\,
\Psi^{(\ell)}(\delta z_{t+1}^{(\ell)}),
\label{eq:hidden_coupling}
\end{equation}
where $\Psi^{(\ell)}$ is a learned linear map and $\alpha_{\mathrm{corr}}$ is a scalar coupling strength. This step closes the predict--correct loop: the latent correction not only affects the latent carried forward, but also nudges the internal memory dynamics so that future priors are generated from a hidden state consistent with the corrected latent trajectory.
Combining Eq.~\eqref{eq:aligned_steps}--\eqref{eq:hidden_coupling} yields a multi-rate latent dynamical model with an explicit predict--correct mechanism. The state predictor consumes recurrent priors (and optionally hidden features) to advance the state, while the latent fusion and hidden coupling act as a stabilizing correction that mitigates drift under long autoregressive rollouts.

\subsection{Training objectives and schedules}\label{sec:trainobj}

We train MSR-HINE by TBPTT on fixed-length windows
$\{u_t\}_{t=0}^{L}$ sampled from each trajectory. At each step $t$, the model produces a one-step
forecast $\hat{u}_{t+1}$ together with multi-level latent priors $\{z^{(\ell)}_{t+1|t}\}_{\ell=1}^{\mathcal{L}}$
and hidden states $\{h^{(\ell)}_{t+1}\}_{\ell=1}^{\mathcal{L}}$.

\paragraph{Per-step forecasting loss.}
For both dynamical systems, we supervise the one-step forecast using a weighted combination of
state-space error and spectral/physics-aware regularizers:
\begin{equation}
\mathcal{L}_{\text{1-step}}
=\sum_{t=0}^{\Ltbptt-1}
\Big(
\lambda_u \,\underbrace{\| \hat{u}_{t+1}-u_{t+1}\|_2^2}_{\mathcal{L}_{u}}
+\lambda_{\text{spec}}\,\mathcal{L}_{\text{spec}}(\hat{u}_{t+1},u_{t+1})
+\lambda_{\text{prior}}\,\mathcal{L}_{\text{prior}}(t)
+\lambda_E\,\mathcal{L}_{E}(\hat{u}_{t+1},u_{t+1})
+\mathcal{L}_{h}(t)
\Big).
\end{equation}
We use $\mathcal{L}_{u}$ as mean-squared error (MSE) in state space.
The low-frequency spectral term $\mathcal{L}_{\text{spec}}$ matches the large-scale Fourier content
(e.g., low-$k$ rFFT bins) and is instantiated slightly differently for KS vs.\ L96 (relative vs.\ absolute,
and whether the DC component is excluded), but follows the same principle: penalize mismatch in
$\Re\{\mathrm{FFT}(\cdot)\}$ and $\Im\{\mathrm{FFT}(\cdot)\}$ at low wavenumbers.

\paragraph{Latent prior matching.}
At each level $\ell$, the recurrent hierarchy predicts a prior $z^{(\ell)}_{t+1|t}$ that should be consistent
with the latent encoding of the \emph{ground-truth} next state,
\begin{equation}
z^{*(\ell)}_{t+1} \;\triangleq\; T^{(\ell)}(u_{t+1}),
\qquad
\mathcal{L}_{\text{prior}}(t) \;\triangleq\; \sum_{\ell=1}^{\mathcal{L}}
d\!\left(z^{(\ell)}_{t+1|t},\,z^{*(\ell)}_{t+1}\right).
\end{equation}
For L96 we take $d(\cdot,\cdot)$ to be MSE in latent space.
For KS, where Fourier coefficients are naturally interpreted in amplitude/phase form, we use a
phase-and-log-amplitude discrepancy (phase emphasized more strongly at coarser levels) to encourage
accurate long-wave structure while stabilizing training.

\paragraph{Spectral energy regularization.}
To further align the multiscale statistics of forecasts, we optionally include a multiband energy term
\begin{equation}
\mathcal{L}_{E}(\hat{u},u)
\;=\;
\sum_{b} w_b \,\Big(\log\frac{E_b(\hat{u})}{E_b(u)}\Big)^{\!2},
\qquad
E_b(u)=\sum_{k\in \mathcal{K}_b} |\widehat{u}(k)|^2,
\end{equation}
where $\mathcal{K}_b$ denotes rFFT index ranges (bands) and $w_b$ are fixed band weights.

\paragraph{Hidden-state drift penalty.}
We regularize the latent memories to evolve smoothly. In the multirate setting, we primarily penalize
\emph{off-stride} updates (i.e., steps where a level is not scheduled for a full update) to discourage
unnecessary hidden drift:
\begin{equation}
\mathcal{L}_{h}(t)
=\gamma_h \sum_{\ell=1}^{\mathcal{L}} \mathbf{1}\{\text{off-stride at }(\ell,t)\}\;
\|h^{(\ell)}_{t+1}-h^{(\ell)}_{t}\|_2^2,
\end{equation}
with $\gamma_h \ge 0$.

\paragraph{Tail-free training within TBPTT windows.}
To reduce exposure bias and improve rollout stability, we split each TBPTT window into a warm prefix
and a free tail. Let $L_{\text{tail}}$ be the number of tail steps and $L_{\text{warm}}=\Ltbptt-L_{\text{tail}}$.
During the warm prefix, the model is trained primarily under teacher forcing, optionally with scheduled
sampling (probability $p_{\text{ss}}$ of feeding back predictions). During the tail, we always feed back
predictions $\hat{u}_{t}$ and form posteriors from predicted states (to mirror inference-time operation).
Crucially, the \emph{prior-matching} target $z^{*(\ell)}_{t+1}=T^{(\ell)}(u_{t+1})$ remains ground-truth
throughout (including the free tail), so the recurrent priors are always trained toward the true latent
future even when the state trajectory is self-fed.

\paragraph{$K$-step emphasis.}
We use two complementary mechanisms to emphasize multi-step accuracy:
(i) \emph{in-tail horizon weighting}, which increases the weight of state error later in the free tail,
\begin{equation}
\mathcal{L}_{\text{tail-wt}}
=\sum_{k=1}^{L_{\text{tail}}} \alpha_k \,\|\hat{u}_{L_{\text{warm}}+k}-u_{L_{\text{warm}}+k}\|_2^2,
\qquad
\alpha_k \propto \tfrac{k}{L_{\text{tail}}},
\end{equation}
and (ii) an optional \emph{explicit $K$-step loss} computed by rolling the model forward an additional
$K$ steps from the end of the TBPTT unroll (fully self-fed) and averaging MSE against the corresponding
ground-truth frames. The latter directly trains the closed-loop predictor on a short inference-style rollout.

\paragraph{Staged schedules (delayed ramps).}
All robustness mechanisms are introduced gradually via delayed linear ramps over epochs. Concretely,
for a generic quantity $v$ (e.g., $p_{\text{ss}}$, $L_{\text{tail}}$, or $\lambda_{\text{prior}}$),
\begin{equation}
\mathrm{ramp}(e; e_0, d, v_{\max})
=
v_{\max}\cdot \mathrm{clip}\!\Big(\frac{e-e_0}{d},\,0,\,1\Big),
\end{equation}
where $e$ is the epoch index, $e_0$ is the start epoch, and $d$ is the ramp duration. We use this form
to (a) ramp scheduled sampling from $0$ to $p_{\text{ss,max}}$, (b) ramp the number of free-tail steps
from $0$ to a chosen fraction of $L$, (c) ramp auxiliary loss weights $\lambda_{\text{prior}}$,
$\lambda_{\text{spec}}$, $\lambda_E$, and (d) ramp the $K$-step loss weight and (optionally) the
$K$ horizon itself. This yields a practical three-stage curriculum: \emph{stabilize} (mostly teacher forcing),
\emph{introduce} (begin mixing self-fed steps and latent/physics losses), and \emph{robustify}
(longer free tails and stronger multi-step emphasis).

\section{Experimental Setup}
\label{sec:exps}

\subsection{Dataset and preprocessing}
\label{sec:exp:data}
For both dynamical systems, we organize data as an array
$\mathbf{X}\in\mathbb{R}^{S\times T\times N}$, where $S$ is the number of independent trajectories,
$T$ is the number of saved time steps per trajectory, and $N$ is the state dimension
($N$ spatial grid points for KS; $N$ variables on the ring for L96).
Training uses \textbf{windowed segments} of length $(\Ltbptt{+}1)$ for TBPTT: each sample is
$\{\mathbf{x}_{t_0},\ldots,\mathbf{x}_{t_0+L}\}\in\mathbb{R}^{(\Ltbptt+1)\times N}$, produced by sliding
a window along time with stride $\Delta t_{\text{win}}$ (``stride'' in the dataloader).

We split along the \emph{trajectory index} $s$ only (entire trajectories are assigned to exactly one split),
so that windows from a given trajectory never appear across multiple splits.

We compute normalization statistics on the \emph{training split only} and apply the same transform to all splits.
We report results in normalized coordinates unless otherwise stated; denormalization is used only for visualization
or to report physical-scale metrics.

For KS, windowed training samples are drawn from normalized trajectories using a sliding window of length $(\Ltbptt{+}1)$.
We also construct \emph{full-trajectory} datasets for evaluation, enabling long free-rollouts from complete sequences.
Unless otherwise stated, our KS experiments use $\Ltbptt=32$, stride $=1$, and batch size $=256$.

For L96, we similarly construct windowed samples of length  $\Ltbptt=16$ and stride $=1$ from normalized trajectories.

We support several normalization modes (global, per-variable, or per-sample) to control whether statistics are shared
across the ring, across trajectories, or both. 
and batch size $=128$.
\subsection{Model configuration}
\label{sec:exp:model}

\subsubsection{Periodic convolutions and residual blocks}
\label{sec:exp:periodic}

Both KS and L96 live on periodic domains (spatially periodic grid for KS; ring topology for L96). We therefore
use circular padding for all 1D convolutions. Given a sequence $x\in\mathbb{R}^{B\times C\times N}$, we apply
a convolutional layer as
\[
\mathrm{PeriodicConv}(x) \;=\; \mathrm{Conv1D}\!\left(\mathrm{pad}_{\mathrm{circ}}(x, p),\, k\right),
\qquad p=\tfrac{k-1}{2},
\]
where $\mathrm{pad}_{\mathrm{circ}}$ performs circular padding along the last dimension.

A residual block at width $C$ consists of two periodic convolutions with GELU nonlinearity and an identity skip:
\[
\mathrm{ResBlock}(x) \;=\; \sigma\!\big(\mathrm{Conv}_2(\sigma(\mathrm{Conv}_1(x))) + x\big),
\]
with $\sigma(\cdot)$ a GELU activation.

\subsubsection{Multiscale 1D U-Net with latent injections}
\label{sec:exp:unet}

Our predictor is a 1D U-Net-style encoder--decoder with three encoder stages and two decoder stages. Let
$\mathbf{u}_t\in\mathbb{R}^{B\times 1\times N}$ be the current state and let
$\mathbf{z}^{(1)}\in\mathbb{R}^{B\times d_1}$ (fine/mid) and $\mathbf{z}^{(2)}\in\mathbb{R}^{B\times d_2}$
(coarse) denote the two injected latent vectors (in practice, priors or concatenations of priors and hidden
summaries, depending on the system).

The encoder produces feature maps at widths $(b,2b,4b)$:
\[
x_1 = E_1(\mathbf{u}_t),\quad
x_2 = E_2(x_1),\quad
x_3 = E_3(x_2),
\]
where each $E_i$ is a periodic convolution followed by a residual block.

We inject $\mathbf{z}^{(1)}$ at the intermediate scale (width $2b$) and $\mathbf{z}^{(2)}$ at the coarsest scale
(width $4b$) via learned MLPs:
\[
\tilde{x}_2 = x_2 + \mathrm{MLP}_2(\mathbf{z}^{(1)})\otimes \mathbf{1}_N,\qquad
\tilde{x}_3 = x_3 + \mathrm{MLP}_3(\mathbf{z}^{(2)})\otimes \mathbf{1}_N,
\]
where $\mathrm{MLP}_2:\mathbb{R}^{d_1}\to\mathbb{R}^{2b}$ and $\mathrm{MLP}_3:\mathbb{R}^{d_2}\to\mathbb{R}^{4b}$
are two-layer SiLU MLPs, and $\otimes \mathbf{1}_N$ denotes broadcasting along the spatial dimension.

The decoder uses skip connections from encoder features:
\[
y_2 = D_2(\tilde{x}_3) + \tilde{x}_2,\qquad
y_1 = D_1(y_2) + x_1,
\]
followed by a periodic $1\times 3$ head to produce an update $\Delta \mathbf{u}_t$:
\[
\Delta \mathbf{u}_t = \alpha\,\tanh(\mathrm{Head}(y_1)),
\]
where $\alpha>0$ is a scalar ``update scale'' (denoted \texttt{u\_scale} in code). In most experiments we predict
a residual update,
\[
\hat{\mathbf{u}}_{t+1} = \mathbf{u}_t + \Delta \mathbf{u}_t,
\]
though the same backbone can be used in a direct-prediction mode.

\subsubsection{Gated fusion module}
\label{sec:exp:gate}

At each level $l\in\{1,2\}$, we fuse a prior latent $\mathbf{z}^{(l)}_{t+1|t}$ with a posterior latent
$\mathbf{z}^{(l)}_{t+1}$ using an elementwise gate:
\[
\mathbf{g}^{(l)}_{t+1} = \sigma\!\left(W^{(l)}\,[\mathbf{z}^{(l)}_{t+1|t};\mathbf{z}^{(l)}_{t+1}] + \mathbf{b}^{(l)}\right),
\]
\[
\mathbf{z}^{(l)}_{t+1,\mathrm{fused}} =
\mathbf{g}^{(l)}_{t+1}\odot \mathbf{z}^{(l)}_{t+1}
+ \bigl(1-\mathbf{g}^{(l)}_{t+1}\bigr)\odot \mathbf{z}^{(l)}_{t+1|t}.
\]
We initialize the gate bias so that $\mathbf{g}^{(l)}\approx p_0$ at initialization, encouraging early reliance on
posterior information during teacher-forced training.

\subsubsection{Per-level recurrent prior with learned latent scaling}
\label{sec:exp:levelrnn}

Each latent level uses a GRU to evolve a hidden state $\mathbf{h}^{(l)}_t\in\mathbb{R}^{d_{\mathrm{hid}}^{(l)}}$
from conditioning features $\mathbf{c}^{(l)}_t\in\mathbb{R}^{d_{\mathrm{in}}^{(l)}}$:
\[
\mathbf{h}^{(l)}_{t+1} = \mathrm{GRU}^{(l)}(\mathbf{c}^{(l)}_t,\mathbf{h}^{(l)}_t).
\]
A linear ``readout'' maps hidden states to latent priors:
\[
\tilde{\mathbf{z}}^{(l)}_{t+1|t} = \tanh\!\left(A^{(l)}\mathbf{h}^{(l)}_{t+1}+\mathbf{a}^{(l)}\right),
\qquad
\mathbf{z}^{(l)}_{t+1|t} = \exp(\mathbf{s}^{(l)})\odot \tilde{\mathbf{z}}^{(l)}_{t+1|t},
\]
where $\mathbf{s}^{(l)}\in\mathbb{R}^{d_{\mathrm{lat}}^{(l)}}$ is a \emph{learned per-latent log-scale}. This
parameterization stabilizes training by allowing the model to adapt latent magnitudes while keeping the raw readout
bounded (via $\tanh$). We orthogonally initialize the recurrent weights to further improve stability.

\paragraph{Injection dimensionality.}
We  concatenate hidden summaries with latents before injection (matching the MSR-HINE design),
so the U-Net injection dimensions become $(d_{\mathrm{lat}}^{(1)}{+}d_{\mathrm{hid}}^{(1)},\,
d_{\mathrm{lat}}^{(2)}{+}d_{\mathrm{hid}}^{(2)})$. 

\subsection{Baselines}
\label{sec:baselines}

We compare MSR-HINE against two autoregressive baselines designed to isolate the impact of (i) multi-rate latent recurrence and (ii) latent hierarchical structure beyond an explicit one-step predictor.

\paragraph{(B1) Two-level HINE.}
Our primary baseline is the two-level hierarchical implicit neural emulator proposed by \cite{jiang2025hierarchicalimplicitneuralemulators} (denoted \textbf{\hinel}). The model augments the current state with a coarse-grained latent representation of a future state, obtained via fixed spatial downsampling, and predicts the next state through an implicit formulation that conditions on this future context. During training, ground-truth future latents are provided, while during inference they are replaced by autoregressive predictions from previous steps. We implement this baseline using the same 1D periodic U-Net backbone and comparable latent dimensionality as MSR-HINE to ensure differences arise from recurrence and hierarchical coupling rather than model capacity.

\begin{equation}
\hat{u}_{n+1}, 
\hat{z}^{(1)}_{n+2}, 
\ldots, 
\hat{z}^{(L-1)}_{n+L}
=
\mathcal{F}_{\theta}
\big(
u_n,
z^{(1)}_{n+1},
\ldots,
z^{(L-1)}_{n+L-1}
\big).
\end{equation}

The operator $\mathcal{F}_\theta$ is a learned neural transition map parameterized by $\theta$, implemented as a 1D periodic U-Net, which jointly predicts the next physical state and a hierarchy of future latent representations. 
The variables $z^{(l)}_{n+k} = \mathcal{T}^{(l)}(u_{n+k})$ denote coarse-grained latent encodings of the state at future time $n+k$, obtained via fixed spatial downsampling operators $\mathcal{T}^{(l)}$, with increasing levels of abstraction indexed by $l = 1, \ldots, L-1$. 
During training, the future latents $\{z^{(l)}_{n+k}\}$ are computed from ground-truth states and provided as conditioning inputs, while during inference they are replaced by the model’s own autoregressive predictions $\{\hat{z}^{(l)}_{n+k}\}$ from previous rollout steps. 
The outputs $\hat{u}_{n+1}$ and $\hat{z}^{(l)}_{n+l+1}$ represent, respectively, the predicted next state and the predicted future latent representations used to condition subsequent predictions in the autoregressive rollout.

\paragraph{(B2) Explicit one-step U-Net predictor (L=1).}
We also consider a simple explicit autoregressive baseline (\textbf{U-Net-AR}) that predicts the next state directly from the current state:
\begin{equation}
\hat{\mathbf{u}}_{t+1} = \mathcal{F}_{\theta}(\mathbf{u}_{t}),
\end{equation}
where $\mathcal{F}_{\theta}$ is a 1D periodic U-Net trained with teacher forcing. This model uses no latent hierarchy and no recurrent memory beyond the autoregressive loop, providing a clean reference for stability and error accumulation under free-rollouts.

\subsection{Training Details}
\label{sec:training_details}

We train all models on fixed-length windows sampled from long trajectories. Each training sample is a chunk of length $\Ltbptt+1$, enabling an unroll of $L$ one-step predictions during training. Optimization is performed with AdamW, gradient clipping, and learning-rate scheduling as described below. 

For KS, we train MSR-HINE for $80$ epochs with AdamW (learning rate $10^{-3}$, weight decay $10^{-4}$), gradient clipping ($0.7$), and a reduce on plateau schedule (patience $6$, factor $0.5$, minimum learning rate $10^{-5}$). We use a TBPTT unroll length of $\Ltbptt=32$. Training employs delayed ramps for (i) scheduled sampling, (ii) free-tail fraction, and (iii) auxiliary losses and horizon-weighted losses. Concretely, scheduled sampling is introduced at epoch $30$ and ramped over $45$ epochs to its maximum probability; the free-tail fraction is introduced at epoch $35$ and ramped over $45$ epochs up to a maximum tail fraction of $0.6$ of the TBPTT window. Auxiliary terms are activated with delayed ramps (e.g., energy-band loss from epoch $20$ over $30$ epochs; horizon-weighted $k$-step emphasis from epoch $18$ over $20$ epochs). During validation, we also ramp a free-tail evaluation regime up to a maximum validation tail fraction of $0.8$ (starting at epoch $5$ over $30$ epochs). 


For L96, we train for $80$ epochs with AdamW (learning rate $10^{-3}$, weight decay $10^{-4}$) and cosine annealing learning-rate schedule. Training uses a TBPTT unroll length $\Ltbptt=16$ and scheduled sampling ramped linearly from $0$ to $0.5$ over $70$ epochs. We optionally introduce a short free tail during training by ramping the number of free-rollout steps from $0$ up to a maximum tail length of 8 beginning at epoch 20 and ramping over 20 epochs. 


\noindent
We checkpoint the best model according to the validation loss computed on teacher-forced steps (with the same regularization terms as in training).

\subsection{Evaluation Protocols and Metrics}
\label{sec:eval_metrics}

All models are evaluated under consistent warm-started autoregressive rollouts on held-out test trajectories. For a given trajectory, we select an initial time index $t_0$, warm-start the model for $W$ steps using teacher forcing to initialize recurrent states and caches, and then perform a free-rollout for $H$ steps by recursively feeding predictions:
\begin{equation}
\hat{\mathbf{u}}_{t+1} = \mathcal{M}_{\theta}(\hat{\mathbf{u}}_{t}, \mathcal{S}_{t}), \qquad
t=t_0+W,\dots,t_0+W+H-1,
\end{equation}
where $\mathcal{S}_{t}$ denotes internal recurrent states (e.g., multiscale hidden states and caches). Metrics are aggregated over multiple trajectories and start times.

\paragraph{State-space error.}
We report MSE/RMSE in state space over the rollout horizon:
\begin{equation}
\mathrm{RMSE}(h) = \left(
\frac{1}{N}\sum_{i=1}^{N} \left(\hat{u}_{t_0+h,i} - u_{t_0+h,i}\right)^2
\right)^{1/2},
\end{equation}
and the time-averaged RMSE over a horizon window. For L96 we also report \emph{forecast RMSE} (FRMSE), computed over the \emph{free-rollout} segment only (excluding warm-start steps).

\paragraph{Spectral fidelity.}
To assess large-scale structure, we report low-wavenumber spectral error based on the mismatch of low-$k$ Fourier coefficients. To diagnose artificial dissipation or spurious energy growth, we additionally evaluate multi-band spectral energy errors computed from the rFFT energy in prescribed bands.

\paragraph{Qualitative diagnostics.}
We include space--time plots (ground truth vs prediction), error heatmaps, and representative single-variable traces (for L96) or spatial snapshots and spectra (for KS). These visualizations highlight common long-horizon failure modes such as phase drift, amplitude collapse, or excessive damping of small scales.

\paragraph{Anomaly correlation coefficient (ACC).}
We report the anomaly correlation coefficient (ACC) as a function of forecast horizon. For each rollout $r$ and time index $t$, let $u_t^{(r)}\in\mathbb{R}^{N}$ be the ground-truth state and $\hat{u}_t^{(r)}\in\mathbb{R}^{N}$ the predicted state (with $N$ grid points for KS or $N=40$ state components for L96). We remove the spatial mean at each time step (i.e., remove the DC mode) to form anomalies
\begin{equation}
a_t^{(r)} = u_t^{(r)} - \frac{1}{N}\sum_{i=1}^{N}u_{t,i}^{(r)}, 
\qquad
\hat{a}_t^{(r)} = \hat{u}_t^{(r)} - \frac{1}{N}\sum_{i=1}^{N}\hat{u}_{t,i}^{(r)}.
\end{equation}
ACC is then computed as the normalized inner product (cosine similarity)
\begin{equation}
\mathrm{ACC}^{(r)}(t)=
\frac{\left\langle \hat{a}^{(r)}_{t},\, a^{(r)}_{t}\right\rangle}
{\left\|\hat{a}^{(r)}_{t}\right\|_2\left\|a^{(r)}_{t}\right\|_2+\varepsilon},
\end{equation}
with $\varepsilon=10^{-12}$ for numerical stability. For each horizon $t=0,\dots,H$, we report the mean and standard deviation across rollouts:
\begin{equation}
\mathrm{ACC}(t)=\frac{1}{R}\sum_{r=1}^{R}\mathrm{ACC}^{(r)}(t),
\qquad
\sigma_{\mathrm{ACC}}(t)=\sqrt{\frac{1}{R}\sum_{r=1}^{R}\left(\mathrm{ACC}^{(r)}(t)-\mathrm{ACC}(t)\right)^2}.
\end{equation}
For numerical safety we clip ACC values to $[-1,1]$.

\section{Results}
\label{sec:results}
\subsection{Long-horizon forecasting accuracy}

We first assess long-horizon \emph{autoregressive} forecasting (self-fed rollouts) for both the 1D KS system and the L96 system. In all cases, models are evaluated in the strict one-step setting ($L=1$): the predicted state at time $t+1$ is fed back as input to predict time $t+2$, and so on. Dataset generation, normalization, and train/validation/test splits are described in the KS and L96 setup subsections (see \S\ref{sec:ks_setup} and \S\ref{sec:setup:l96}).

\begin{figure}
    \centering
    \includegraphics[width=\linewidth]{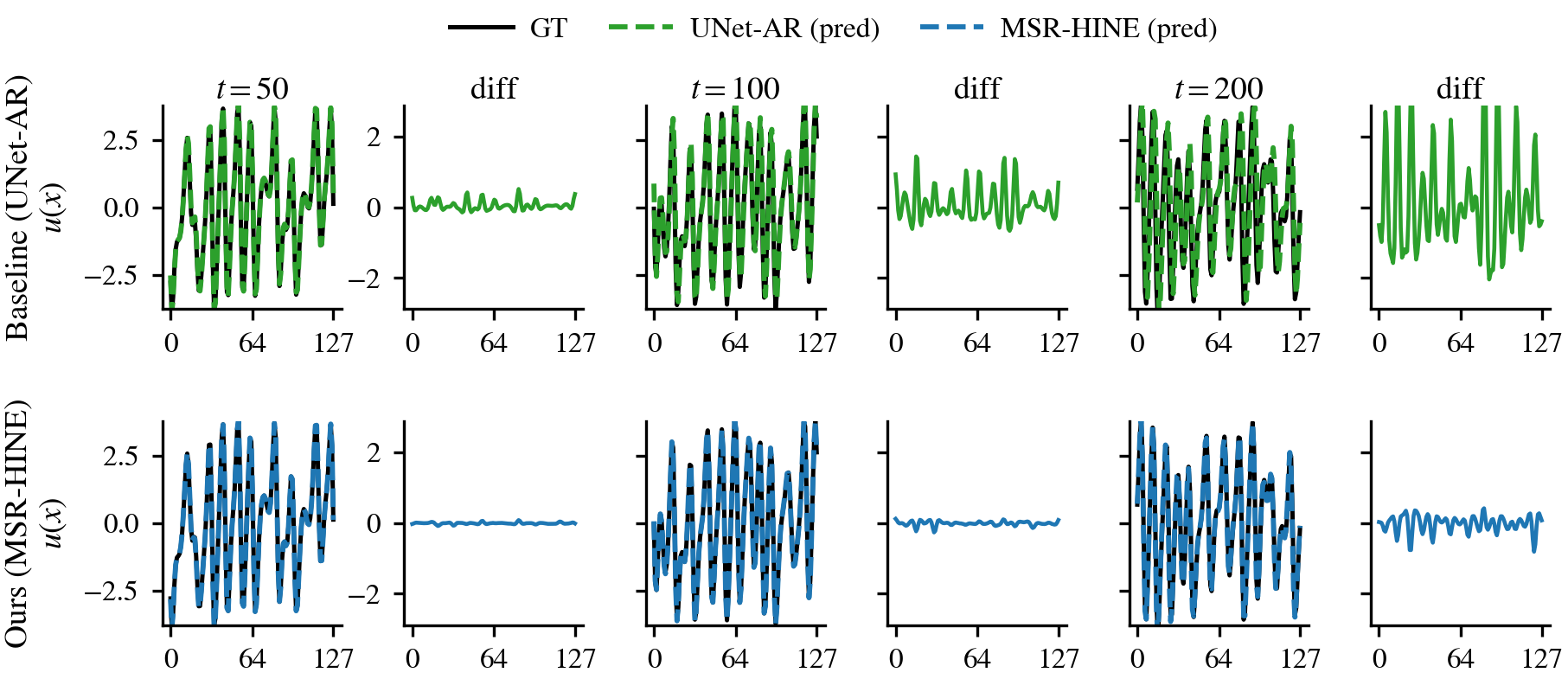}
    \caption{\textbf{Autoregressive KS state profiles (L=1): U-Net-AR vs.\ MSR-HINE.}
  Representative spatial profiles $u(x)$ at increasing rollout horizons (columns). Solid black: ground truth (GT).
  Green dashed: U-Net-AR prediction. Blue dashed: MSR-HINE prediction. The ``diff'' panels report the pointwise error
  (prediction $-$ GT), highlighting stronger error growth and drift for U-Net-AR at longer horizons, while MSR-HINE
  remains more stable under self-fed rollouts.} 
    \label{fig:ks_autoreg}
\end{figure}

Figure~\ref{fig:ks_autoreg} compares representative KS spatial profiles $u(x)$ at increasing horizons for the U-Net autoregressive baseline (U-Net-AR) and MSR-HINE. At short lead times, both approaches match the ground truth closely, indicating comparable one-step predictive skill. As the rollout horizon increases, the U-Net-AR baseline accumulates noticeable phase and amplitude errors, which appear as growing pointwise residuals in the corresponding ``diff'' panels. In contrast, MSR-HINE maintains substantially better alignment with the ground-truth profile at the same horizons, with smaller residuals and reduced drift, suggesting improved rollout stability under chaotic dynamics.

\begin{figure}
    \centering
    \includegraphics[width=\linewidth]{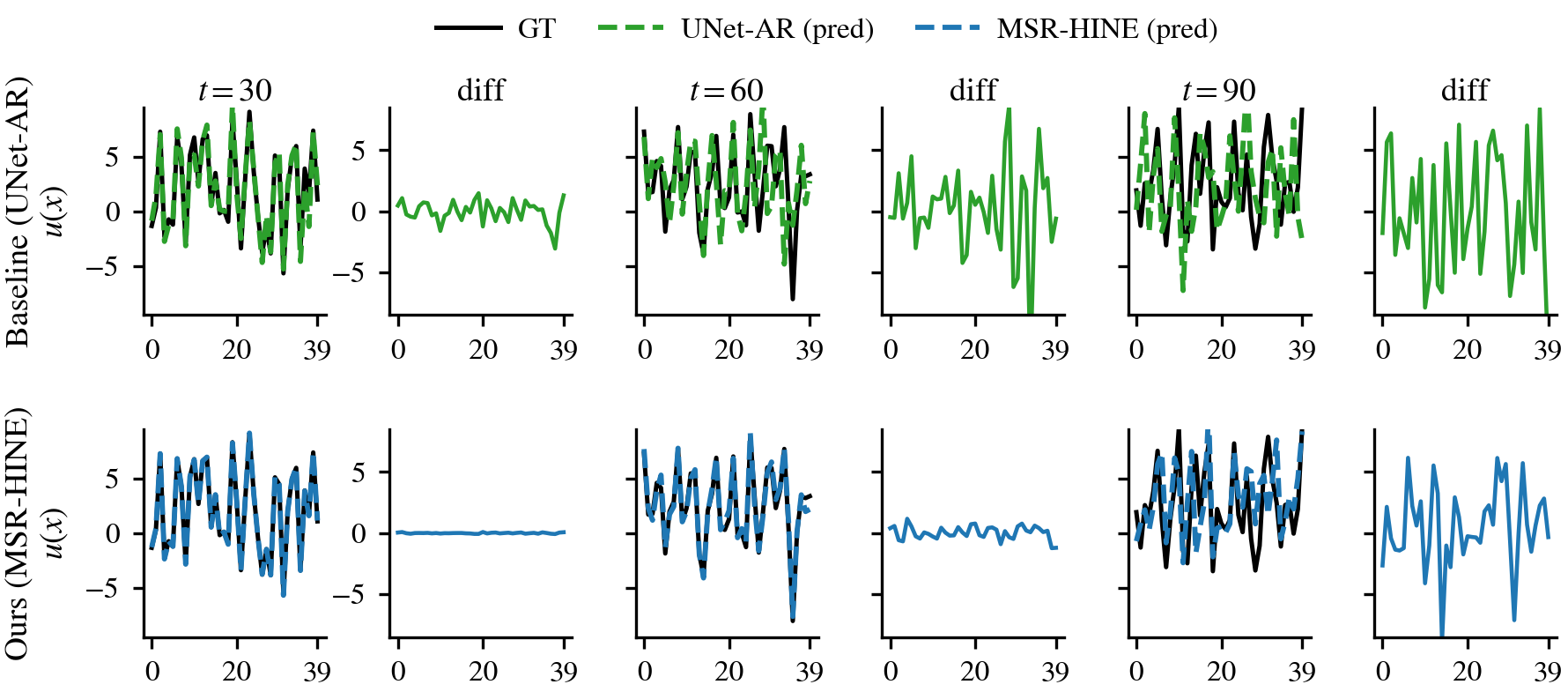}
    \caption{\textbf{Autoregressive L96 state profiles (L=1): U-Net-AR vs.\ MSR-HINE.}
  Representative L96 state vectors $u(x)$ (with $N=40$ cyclic coordinates) shown at horizons $t=\{30,60,90\}$.
  Solid black: ground truth (GT). Green dashed: U-Net-AR prediction. Blue dashed: MSR-HINE prediction.
  The ``diff'' panels show pointwise errors (prediction $-$ GT). Compared to U-Net-AR, MSR-HINE exhibits reduced drift
  and slower error growth over the rollout, indicating improved stability in long-horizon autoregressive forecasting.}
    \label{fig:l96_autoreg}
\end{figure}

A similar trend is observed for L96 in Fig.~\ref{fig:l96_autoreg}, where we plot the state vector $u(x)$ over the $N=40$ cyclic coordinates. While both models track the ground truth well at early times (e.g., $t=30$), the U-Net-AR baseline exhibits rapidly increasing deviations by intermediate horizons (e.g., $t=60$) and pronounced mismatch by later horizons (e.g., $t=90$). MSR-HINE remains closer to the ground truth across these horizons, and its error signal grows more gradually. Overall, these qualitative comparisons indicate that incorporating hierarchical multiscale recurrent latent dynamics improves robustness to error accumulation in autoregressive rollouts relative to a purely feed-forward U-Net baseline.

To complement the qualitative autoregressive profile comparisons in Figs.~\ref{fig:ks_autoreg}--\ref{fig:l96_autoreg}, we report quantitative rollout RMSE at selected lead times for both datasets in Tables~\ref{tab:ks_rmse_horizons} and \ref{tab:l96_rmse_horizons} (mean $\pm$ standard deviation over test rollouts). For KS (Table~\ref{tab:ks_rmse_horizons}), MSR-HINE achieves the lowest error consistently across all horizons, with a clear separation from the U-Net-AR baseline as the rollout length increases (e.g., $t{=}200$: $2.29\times10^{-1}$ vs.\ $1.48$, and $t{=}400$: $1.09$ vs.\ $2.94$). The two-level Jiang-style HINE baseline (\hinel) improves upon U-Net-AR at intermediate and long horizons, but remains uniformly worse than MSR-HINE, indicating that explicitly propagating multiscale recurrent latent dynamics yields a more robust long-horizon emulator in this chaotic regime. 

For L96 (Table~\ref{tab:l96_rmse_horizons}), the same ordering is observed at early and intermediate horizons, where MSR-HINE provides substantial gains over both \hinel and U-Net-AR (e.g., $t{=}10$ and $t{=}25$). At longer horizons ($t\ge 50$), all models experience pronounced error growth, reflecting the strong sensitivity and mixing inherent to L96; nevertheless, MSR-HINE maintains the best mean performance among the three methods at each reported horizon, with reduced variability in the short-horizon regime and competitive stability as the rollout length increases. Together, these results corroborate the profile-level observations: MSR-HINE delays error accumulation and improves rollout fidelity relative to both a purely feed-forward U-Net autoregressive baseline and the hierarchical-implicit HINE variant.

\begin{table}
\centering
\resizebox{\columnwidth}{!}{%
\begin{tabular}{lcccccc}
\toprule
Model & $t{=}10$ & $t{=}25$ & $t{=}50$ & $t{=}100$ & $t{=}200$ & $t{=}400$ \\
\midrule
MSR-HINE & 3.71e-03$\pm$2.95e-04 & 8.40e-03$\pm$1.32e-03 & 2.14e-02$\pm$4.55e-03 & 6.35e-02$\pm$1.41e-02 & 2.29e-01$\pm$7.58e-02 & 1.09e+00$\pm$3.53e-01 \\
HINE-L2  & 1.21e-02$\pm$1.62e-03 & 3.10e-02$\pm$5.00e-03 & 7.71e-02$\pm$1.78e-02 & 2.05e-01$\pm$4.96e-02 & 7.14e-01$\pm$3.44e-01 & 1.92e+00$\pm$4.11e-01 \\
U-Net-AR  & 2.16e-02$\pm$3.35e-03 & 5.88e-02$\pm$6.47e-03 & 1.57e-01$\pm$1.52e-02 & 4.62e-01$\pm$4.98e-02 & 1.48e+00$\pm$1.08e-01 & 2.94e+00$\pm$1.77e-01 \\
\bottomrule
\end{tabular}%
}
\caption{\textbf{KS autoregressive rollout RMSE at selected horizons.}
Mean $\pm$ standard deviation over test trajectories for one-step autoregressive rollouts (self-fed predictions).
Lower is better. MSR-HINE consistently achieves the lowest RMSE and exhibits slower error growth with horizon relative
to both HINE-L2 and the U-Net-AR baseline.}
\label{tab:ks_rmse_horizons}
\end{table}

\begin{table}
\centering
\resizebox{\columnwidth}{!}{%
\begin{tabular}{lccccc}
\toprule
Model & t=10 & t=25 & t=50 & t=75 & t=100 \\
\midrule
MSR-HINE & 3.71e-03±2.95e-04 & 8.40e-03±1.32e-03 & 2.62e-01±2.22e-01 & 1.62e+00±9.15e-01 & 3.44e+00±8.00e-01 \\
HINE-L2 & 7.12e-02±1.65e-02 & 4.29e-01±2.38e-01 & 2.94e+00±1.14e+00 & 4.49e+00±7.58e-01 & 4.98e+00±5.85e-01 \\
U-Net-AR & 8.78e-02±2.98e-02 & 6.07e-01±4.01e-01 & 2.86e+00±1.26e+00 & 4.40e+00±7.54e-01 & 4.72e+00±4.63e-01 \\
\bottomrule
\end{tabular}%
}
\caption{\textbf{L96 autoregressive rollout RMSE at selected horizons.}
Mean $\pm$ standard deviation over test trajectories for one-step autoregressive rollouts.
Lower is better. MSR-HINE yields the best short- to mid-horizon accuracy and remains competitive at longer horizons
where all models degrade due to the strongly chaotic dynamics.}
\label{tab:l96_rmse_horizons}
\end{table}

\begin{figure}
    \centering
    \begin{minipage}{0.48\textwidth}
        \centering  
        \includegraphics[width=\linewidth]{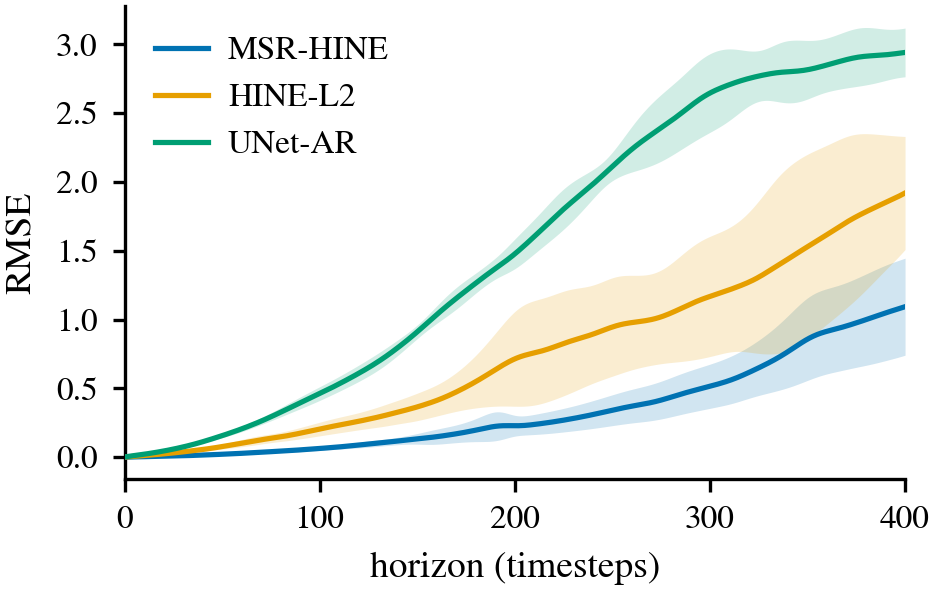}
        \label{fig:rmse}
    \end{minipage}
    \hfill
    \begin{minipage}{0.48\textwidth}
        \centering
        \includegraphics[width=\linewidth]{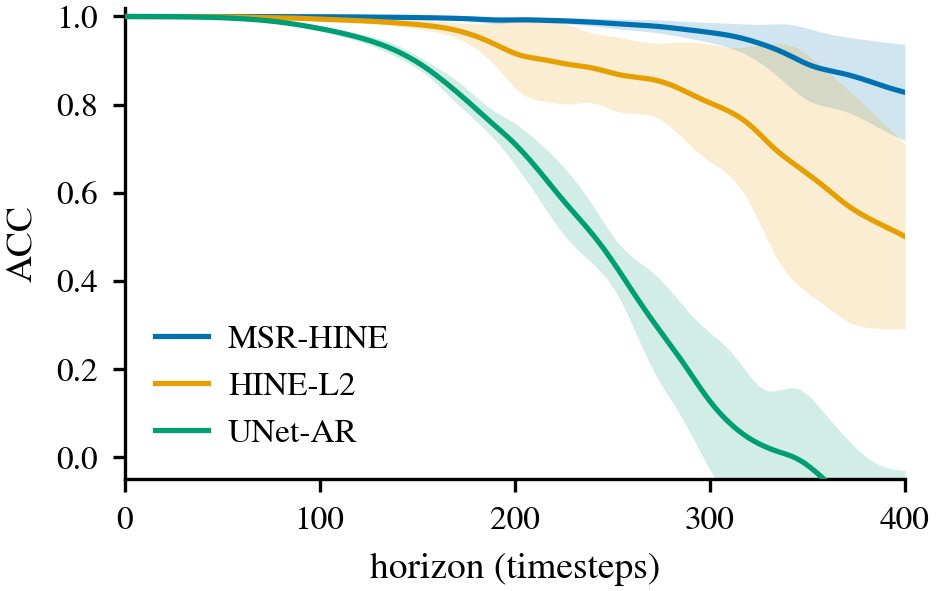}
        \label{fig:acc}
    \end{minipage}
    \caption{Kuramoto--Sivashinsky (KS) long-horizon forecast skill under autoregressive rollouts. Left: RMSE versus horizon. Right: anomaly correlation coefficient (ACC) versus horizon, computed as the cosine similarity between predicted and ground-truth anomalies (space-mean removed at each time). Solid lines denote the mean over test rollouts and shaded bands indicate $\pm$1 standard deviation. MSR-HINE exhibits slower error growth and maintains higher anomaly correlation over longer horizons compared with HINE-L2 and the U-Net-AR baseline.}

    \label{fig:ks_rmse_acc}
\end{figure}

Figure~\ref{fig:ks_rmse_acc} summarizes long-horizon skill on the KS dataset using RMSE and ACC as a function of forecast horizon. MSR-HINE consistently maintains the lowest RMSE across the entire rollout and exhibits the slowest growth rate with horizon, indicating improved stability under autoregressive self-feeding. This behavior is mirrored by the ACC curve: MSR-HINE retains near-perfect anomaly correlation for substantially longer horizons before gradually decaying, whereas both HINE-L2 and the U-Net-AR baseline lose correlation earlier and with larger inter-rollout variability (shaded bands). Notably, the baseline U-Net-AR shows the fastest deterioration in both metrics—RMSE rises sharply and ACC collapses toward zero—suggesting rapid phase drift and pattern decorrelation in the chaotic regime. HINE-L2 improves over U-Net-AR but still exhibits earlier loss of skill than MSR-HINE, consistent with the idea that multiscale recurrent latent propagation provides an additional mechanism for preserving long-term temporal structure beyond single-step autoregression.

\begin{figure}
    \centering
    \begin{minipage}{0.48\textwidth}
        \centering
        \includegraphics[width=\linewidth]{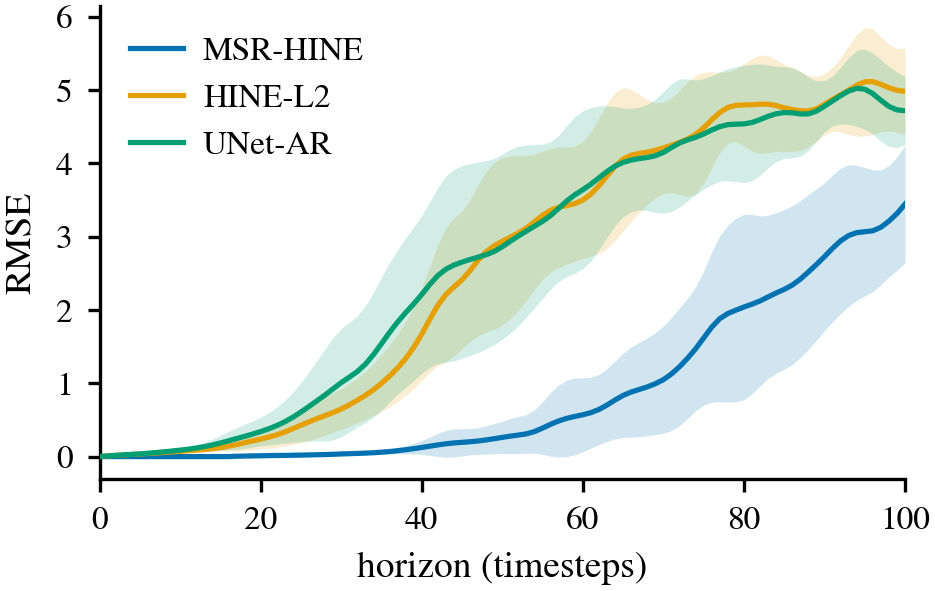}
        \label{fig:rmse}
    \end{minipage}
    \hfill
    \begin{minipage}{0.48\textwidth}
        \centering
        \includegraphics[width=\linewidth]{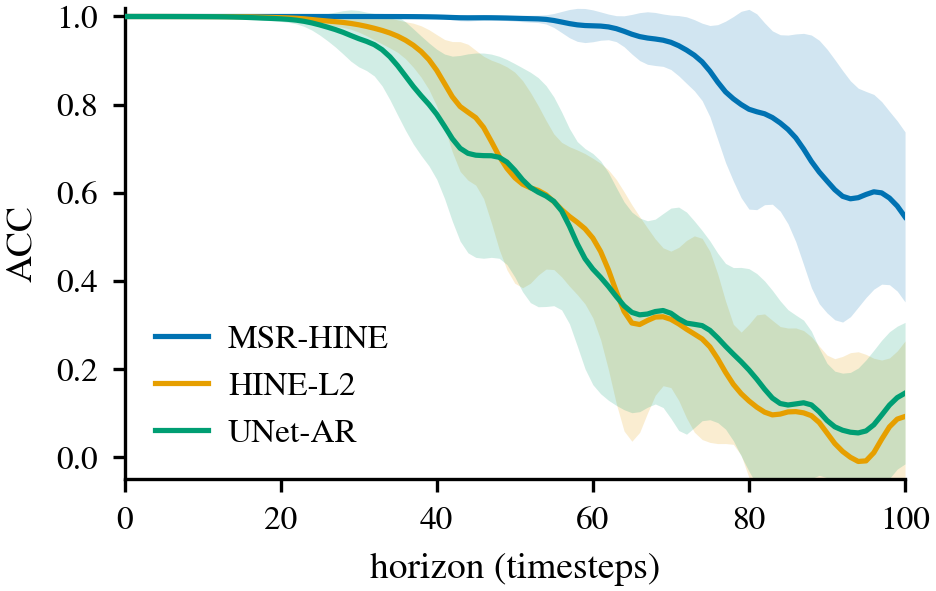}
        \label{fig:acc}
    \end{minipage}
    \caption{L96 long-horizon forecast skill under autoregressive rollouts. Left: RMSE versus horizon. Right: anomaly correlation coefficient (ACC) versus horizon, computed as the cosine similarity between predicted and ground-truth anomalies (space-mean removed at each time). Solid lines denote the mean over test rollouts and shaded bands indicate $\pm$1 standard deviation. Consistent with the more strongly chaotic dynamics, all methods decorrelate more rapidly than in KS; however, MSR-HINE sustains lower RMSE and higher ACC across the rollout relative to HINE-L2 and the U-Net-AR baseline.}

    \label{fig:l96_rmse_acc}
\end{figure}

A similar trend is observed for the L96 dataset in Figure~\ref{fig:l96_rmse_acc}, but with a noticeably shorter predictability window overall, consistent with the strongly chaotic dynamics at the chosen forcing and state dimension (see the L96 setup subsection). MSR-HINE again yields the best long-horizon performance: its RMSE remains significantly lower than the other methods throughout the rollout, and its ACC stays close to one for longer horizons before decaying. In contrast, HINE-L2 and U-Net-AR experience much earlier correlation loss, with ACC dropping rapidly and RMSE saturating at large values; these trends reflect a quicker departure from the correct anomaly subspace and highlight the difficulty of maintaining coherent trajectory evolution in L96 under pure autoregressive rollouts. Importantly, the widening uncertainty bands for HINE-L2 and U-Net-AR suggest sensitivity to initial conditions and stronger rollout-to-rollout variability, while MSR-HINE exhibits more consistent behavior across test trajectories.

Together, these results reinforce the qualitative rollout comparisons: the multiscale recurrent hierarchy in MSR-HINE improves robustness under compounding error, delaying both amplitude blow-up (RMSE) and phase/pattern decorrelation (ACC). This motivates the next analysis, where we directly compare predictability limits using RMSE/ACC thresholds and discuss how the three models trade off early-horizon accuracy versus long-horizon stability across KS and L96.

\subsection{Qualitative rollouts}

\begin{figure}
    \centering
    \includegraphics[width=\linewidth]{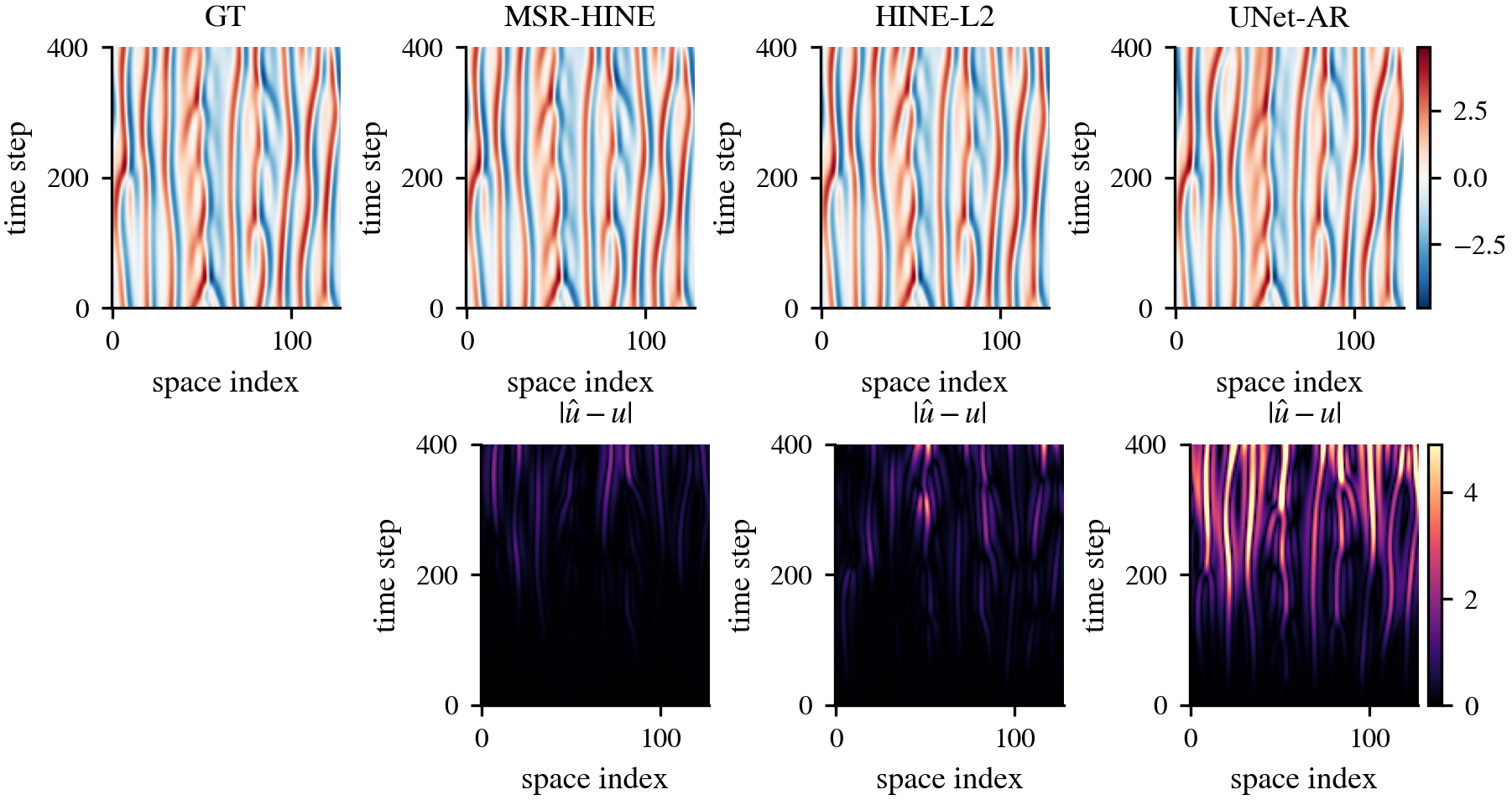}
      \caption{Space--time ground truth $u(x,t)$ (left) and free-rollout predictions from MSR-HINE, HINE-L2, and U-Net-AR over the same horizon (top row), with absolute error maps $|\hat{u}-u|$ for each model (bottom row). MSR-HINE preserves coherent structures and maintains low, localized error; HINE-L2 shows earlier phase drift; U-Net-AR accumulates the largest long-horizon error.}

    \label{fig:ks_rmse_horizons_best}
\end{figure}

\begin{figure}
    \centering
    \includegraphics[width=\linewidth]{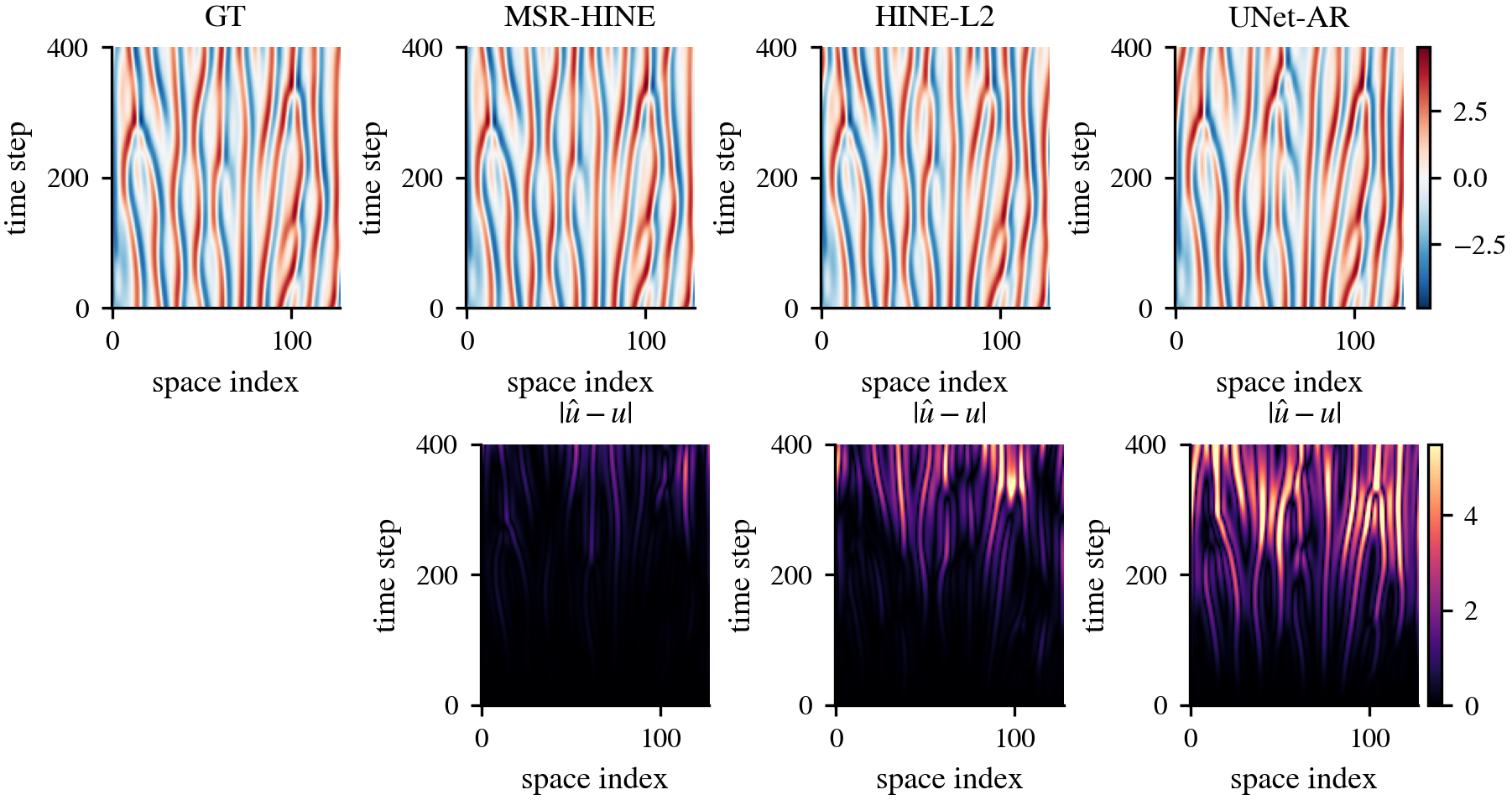}
      \caption{Same layout as the best-case example, but for the most challenging test trajectory (highest rollout error). U-Net-AR diverges earliest with widespread high error, HINE-L2 exhibits intermittent growth and increasing phase mismatch, and MSR-HINE delays global degradation while better retaining large-scale organization, with failure dominated by late-time phase drift.}

    \label{fig:ks_rmse_horizons_worst}
\end{figure}

To complement the quantitative RMSE/ACC analysis, we visualize representative \emph{best-case} in Figure~\ref{fig:ks_rmse_horizons_best} and Figure~\ref{fig:ks_rmse_horizons_worst} \emph{worst-case} KS test trajectories under free (self-fed) autoregressive rollouts. In the best-performing example, all models track the dominant space--time structures at early horizons, but their long-horizon behavior separates. \textbf{MSR-HINE} preserves coherent wave patterns and drift with only mild phase error, remaining visually close to the ground truth for most of the rollout. Accordingly, the corresponding error map $|\hat{u}-u|$ stays dark and spatially localized, indicating that errors are small and do not spread across the domain. \textbf{HINE-L2} remains stable but exhibits earlier phase slippage: the main structures persist while their alignment gradually drifts, producing thicker error streaks over time. \textbf{U-Net-AR} is most sensitive to autoregressive accumulation; even when the early-time match is good, small one-step biases compound, leading to broader late-time error regions and degraded structure.

In the worst-performing example, the chaotic amplification of small inaccuracies is evident. \textbf{U-Net-AR} loses phase coherence rapidly and develops spatially widespread error bands, consistent with a rollout that has diverged from the correct attractor neighborhood. \textbf{HINE-L2} performs better than U-Net-AR but still shows intermittent error bursts and progressively increasing phase mismatch. \textbf{MSR-HINE} remains the most resilient: while errors eventually accumulate, it better preserves the large-scale organization of the space--time field and delays the onset of global divergence. Overall, these best/worst cases reinforce the key takeaway for KS: multiscale recurrent latent evolution stabilizes self-fed rollouts and typically fails via gradual phase drift rather than abrupt structural breakdown.

To complement the aggregate RMSE/ACC statistics reported in Section~\ref{sec:setup:l96}, we visualize representative \emph{best-} and \emph{worst-case} autoregressive rollouts on the Lorenz--96 test set. In both figures, the top row shows space--time diagrams of the state evolution (state index vs.\ time), while the bottom row reports the corresponding pointwise absolute error magnitude, $|\hat{u}-u|$, for each model. The showcased ``best'' and ``worst'' trajectories are selected by the smallest and largest time-averaged rollout error over the fixed evaluation horizon, respectively.

\begin{figure}
    \centering
    \includegraphics[width=\linewidth]{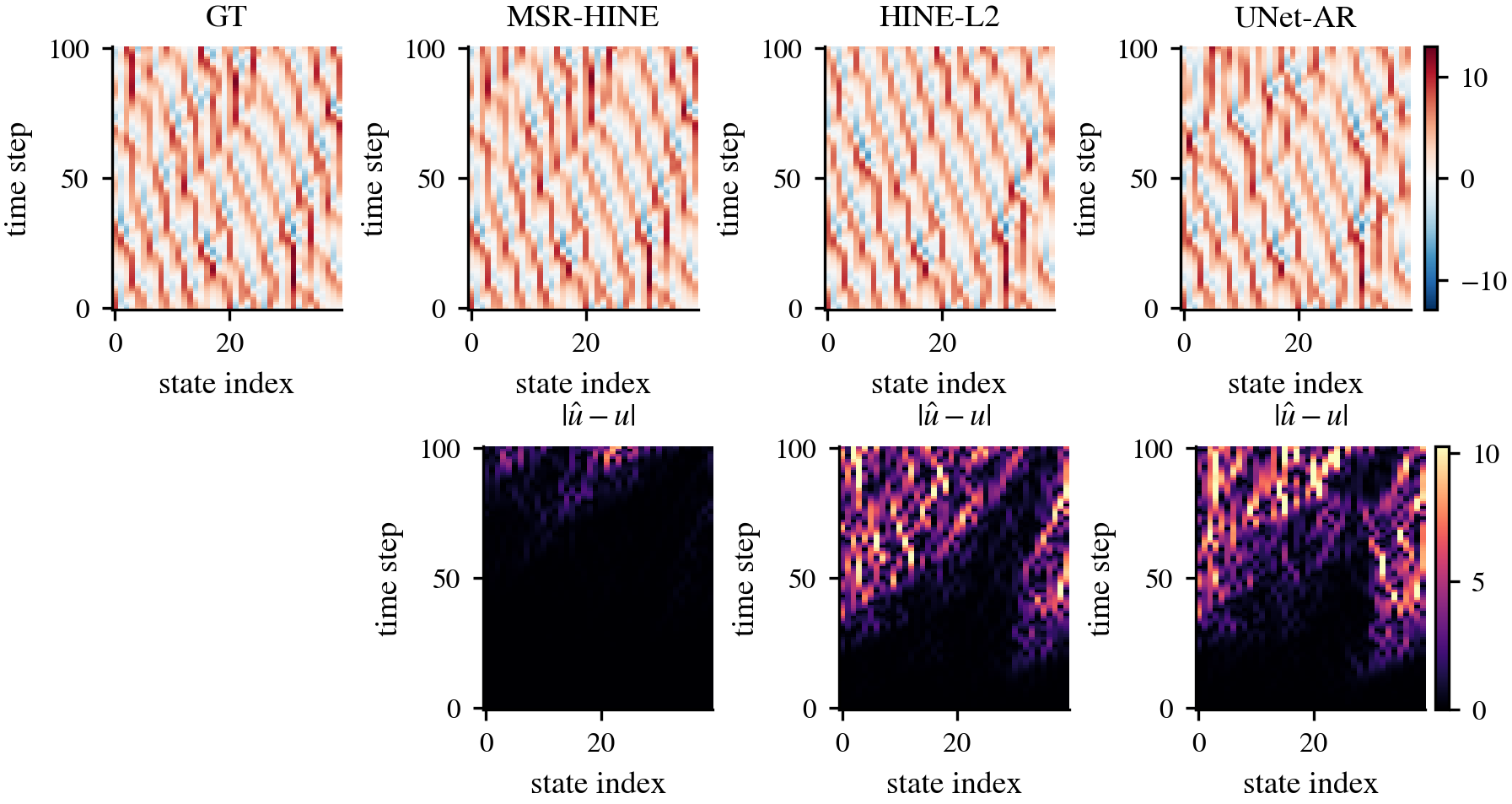}
    \caption{\textbf{Lorenz--96 best-case qualitative rollout (autoregressive forecasting).}
  Space--time diagrams for a representative test trajectory over the evaluation horizon (top row: ground truth and model predictions; bottom row: absolute error $|\hat{u}-u|$ for each method).
  MSR-HINE remains most phase-aligned with the reference and exhibits the smallest, most localized errors, while HINE-L2 and U-Net-AR show earlier and more widespread deviations.}
    \label{fig:l96_rmse_horizons_best}
\end{figure}

Figure~\ref{fig:l96_rmse_horizons_best} highlights a test rollout in which all methods remain qualitatively stable over short horizons, but MSR-HINE most faithfully tracks the ground-truth spatiotemporal pattern. In particular, the predicted space--time bands remain visually coherent and phase-aligned with the reference, and the error map stays close to zero over the majority of the rollout window. In contrast, HINE-L2 and the U-Net-AR baseline exhibit earlier local phase mismatches and intermittent amplitude deviations, visible as localized high-error patches that emerge sooner and occupy a larger fraction of the space--time domain.

\begin{figure}
    \centering
    \includegraphics[width=\linewidth]{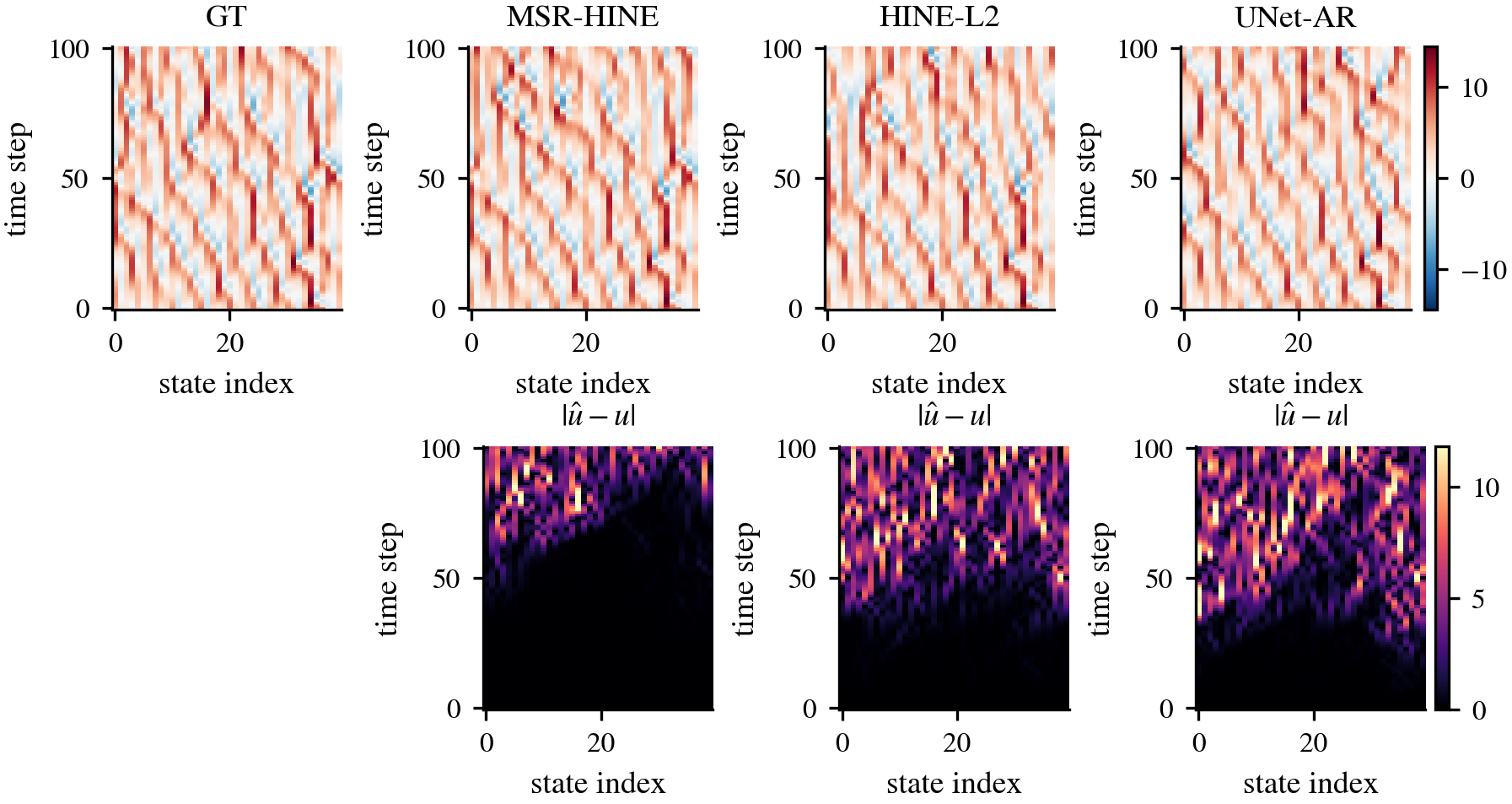}
    \caption{\textbf{Lorenz--96 worst-case qualitative rollout (autoregressive forecasting).}
  Same visualization as Fig.~\ref{fig:l96_rmse_horizons_best}, but for a challenging test trajectory with the largest time-averaged rollout error over the evaluation horizon.
  All methods degrade as the rollout progresses, but MSR-HINE delays the growth of large errors and better preserves coherent spatiotemporal structure relative to HINE-L2 and U-Net-AR.}
    \label{fig:l96_rmse_horizons_worst}
\end{figure}

Figure~\ref{fig:l96_rmse_horizons_worst} shows a challenging test sample where small early discrepancies amplify rapidly due to the intrinsic instability of the L96 dynamics (see Section~\ref{sec:setup:l96}). Here, all models eventually incur noticeable errors; however, MSR-HINE still delays the onset of large-magnitude errors and maintains more recognizable large-scale structure for longer. HINE-L2 and U-Net-AR display broader, more spatially diffuse error regions that appear earlier in time, indicating faster loss of phase coherence and reduced temporal consistency in the free-running regime.

\subsection{Spectral and energy fidelity}

\begin{figure}
    \centering
    \includegraphics[width=\linewidth]{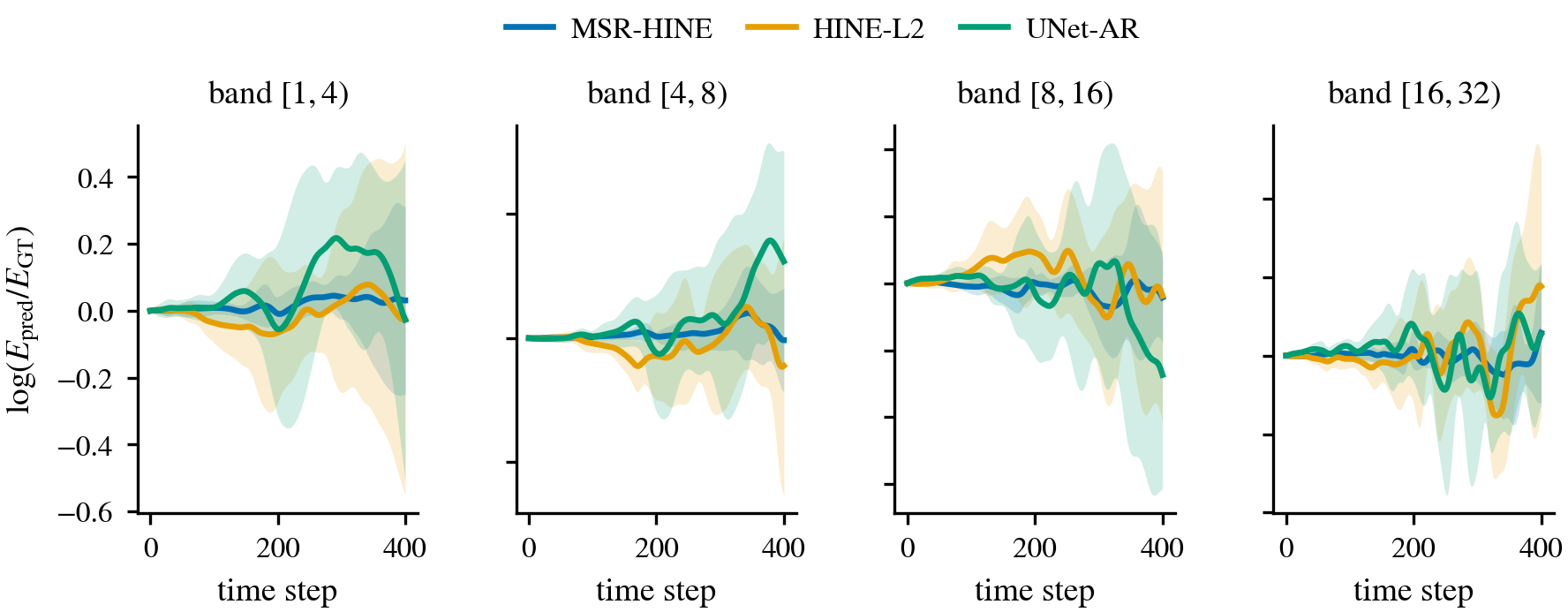}
    \caption{\textbf{KS spectral-band energy fidelity.}
  Evolution of the log band-energy ratio $\log(E_{\mathrm{pred}}/E_{\mathrm{GT}})$ during $400$-step free-rollouts for four Fourier bands: $[1,4)$, $[4,8)$, $[8,16)$, and $[16,32)$.
  Solid lines denote the mean over test rollouts and the shaded regions denote $\pm 1$ standard deviation.
  Values near $0$ indicate correct energy content in that band; negative/positive values indicate under-/over-estimation of band energy.}
    \label{fig:ks_energy_fidelity}
\end{figure}

\begin{figure}
    \centering
    \includegraphics[width=\linewidth]{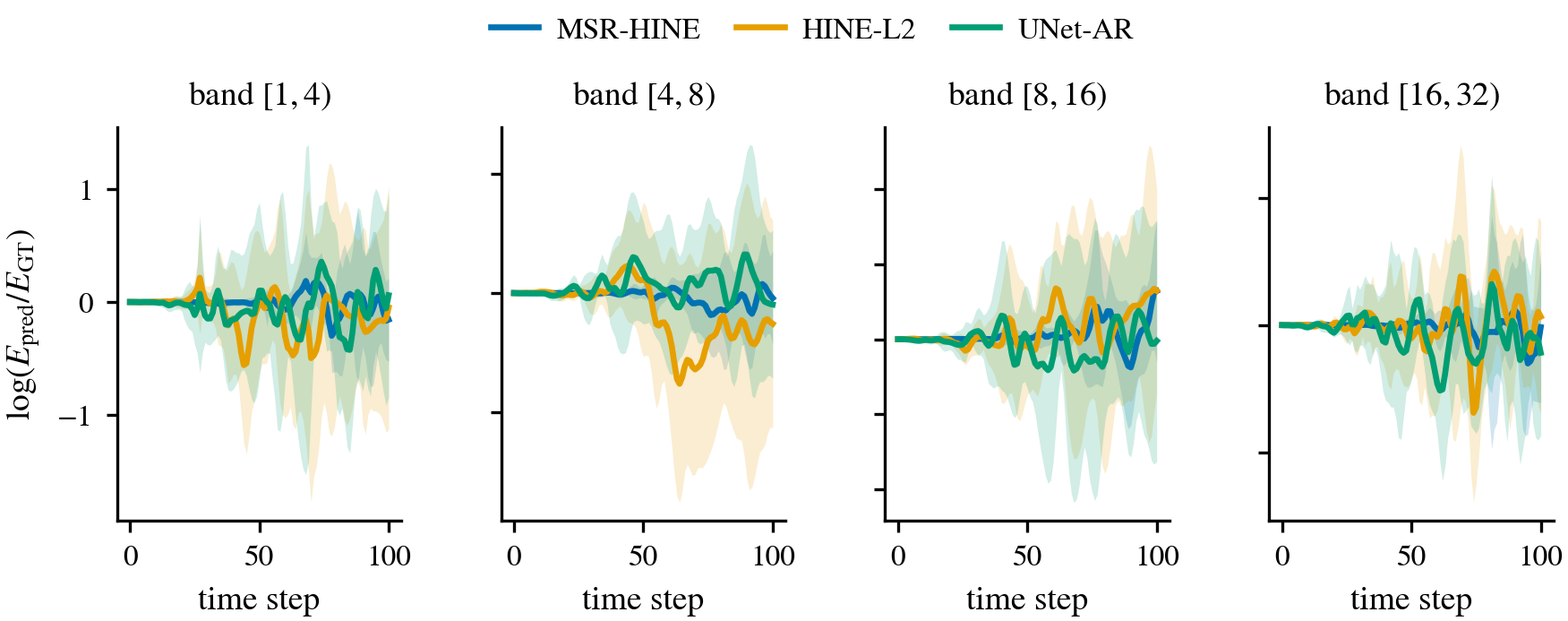}
    \caption{\textbf{L96 spectral-band energy fidelity.}
  Same diagnostic as Fig.~\ref{fig:ks_energy_fidelity}, shown for L96 rollouts (horizon $100$) over bands $[1,4)$, $[4,8)$, $[8,16)$, and $[16,32)$.
  Mean (solid) and $\pm 1$ standard deviation (shaded) are computed over test rollouts.}
    \label{fig:l96_energy_fidelity}
\end{figure}

To assess whether each surrogate preserves the distribution of energy across spatial scales (beyond pointwise errors), we track the evolution of \emph{band-limited} spectral energy during free-rollouts. Specifically, at each time step we compute the energy in Fourier band $[k_1,k_2)$ and report the log-ratio
\begin{equation}
\label{eq:band_energy_ratio}
\log\!\left(\frac{E_{\text{pred}}^{[k_1,k_2)}(t)}{E_{\text{GT}}^{[k_1,k_2)}(t)}\right),
\qquad
E^{[k_1,k_2)}(t) \;=\; \sum_{k=k_1}^{k_2-1} |\hat{u}(k,t)|^2,
\end{equation}
where $\hat{u}(k,t)$ denotes the discrete Fourier coefficients (computed along the spatial/state index). Values near zero indicate accurate energy content in that band, while negative/positive values correspond to under-/over-estimation, respectively. The shaded envelopes in Fig.~\ref{fig:ks_energy_fidelity} and Fig.~\ref{fig:l96_energy_fidelity} show variability over the test rollouts.
For KS, MSR-HINE consistently stays closest to zero across all bands throughout the $400$-step horizon, indicating that it best preserves the multiscale energy distribution over long rollouts. In contrast, U-Net-AR exhibits progressively larger drift and spread across bands, with clear long-horizon bias and amplified uncertainty—consistent with the qualitative degradation observed in long rollouts. HINE-L2 improves over U-Net-AR in several bands but still develops noticeable bias (and increased variance) at longer horizons, especially as the rollout becomes dominated by chaotic error growth. Overall, MSR-HINE’s improved long-horizon stability is accompanied by markedly better spectral-energy consistency, suggesting that multi-rate latent recurrence helps maintain physically meaningful scale interactions.
For L96 (shorter horizon), all models begin near zero and then show increasing dispersion as errors accumulate. MSR-HINE again remains the most tightly clustered around zero across the low and mid wavenumber bands, with smaller long-horizon bias than both HINE-L2 and U-Net-AR. HINE-L2 shows pronounced band-dependent deviations (including sustained under-/over-estimation in intermediate bands) and larger rollout-to-rollout variability, while U-Net-AR displays the broadest spread and intermittent band-energy excursions. These trends align with the RMSE/ACC deterioration at longer horizons: once a model begins misallocating energy across bands, it loses phase coherence and the error growth accelerates.

\section{Discussion}
\label{sec:discussion}
\subsection{Why KS vs L96 differ in difficulty}

Although both Kuramoto--Sivashinsky (KS) and L96 exhibit sensitive dependence on initial conditions, they stress long-horizon forecasters in qualitatively different ways. KS is a spatially-extended PDE with translation-invariant structure and a broad inertial range; errors can manifest as local phase shifts and gradual spectral drift, often remaining visually plausible for some time even when pointwise RMSE grows. In contrast, L96 is a low-dimensional but strongly mixing dynamical system in which global coupling and rapid instability cause forecast errors to amplify into state-space divergence that quickly decorrelates from the ground truth. As a result, correlation-based skill (ACC) typically collapses earlier on L96 than on KS for comparable one-step accuracy, since small phase mismatches in KS may preserve spatial anomalies while L96 errors scramble the anomaly pattern across variables. These dataset-specific dynamics also interact with model inductive bias: convolutional U-Nets can exploit KS locality and periodicity more directly, whereas L96 lacks a true spatial metric and is better viewed as a coupled ring of variables, making long-horizon stability more dependent on accurately maintaining slow-manifold statistics rather than local pattern continuation.

\subsection{Limitations and future work}

Several limitations should be noted. First, all models in this study are trained and evaluated under a fixed normalization scheme and a fixed training horizon (TBPTT window), which may bias performance toward the statistics of the training distribution and can still leave models vulnerable to distribution shift under very long rollouts. Second, our evaluation focuses on autoregressive forecasting from clean initial conditions; robustness to perturbed initial states, partial observations, and observation noise was not systematically explored. Third, the spectral/energy fidelity metric is a compact proxy for physical consistency but does not fully characterize higher-order statistics (e.g., structure functions, intermittency, or higher-order moments), nor does it guarantee accurate reproduction of rare events. Finally, MSR-HINE introduces additional architectural components (multi-rate recurrence, latent fusion, hidden-state correction), which increase implementation complexity and introduce hyperparameters (e.g., stride choices, latent sizes) that require tuning and may be dataset-dependent.

A natural next step is to extend MSR-HINE to \emph{partially observed} settings by coupling the latent hierarchy with data assimilation, where the posterior fusion can incorporate observation-informed latents rather than posteriors derived only from the predicted state. This would enable evaluating robustness under noisy sensors, sparse measurements, and nontrivial observation operators. Second, we plan to investigate adaptive temporal scale selection (learned or data-driven stride scheduling) so that the multi-rate modules can automatically allocate capacity to slow and fast modes depending on regime. Third, improving physical fidelity beyond second-order statistics is an important direction; incorporating explicit constraints or auxiliary losses targeting higher-order moments and scale-to-scale transfer could better preserve long-time invariant measures. Finally, scaling to more realistic PDE systems (e.g., 2D turbulence variants, geophysical surrogates) and testing cross-regime generalization would clarify how multi-rate latent priors and fusion mechanisms transfer to heterogeneous, real-world distributions.

\section{Conclusion}
\label{sec:conclusion}

We presented \emph{MSR-HINE}, a multiscale hierarchical implicit forecaster that augments one-step implicit prediction with \emph{multi-rate recurrent priors} and a \emph{gated posterior fusion} mechanism to stabilize long-horizon rollouts in chaotic systems. Across both the Kuramoto--Sivashinsky (KS) and L96 benchmarks, MSR-HINE consistently improves accuracy, correlation skill, and spectral/energy fidelity relative to a strong U-Net autoregressive baseline and a non-multi-rate hierarchical variant (HINE-L2). 

Quantitatively (from table \ref{tab:conclusion_summary}), at long horizons MSR-HINE yields substantial error reductions and higher anomaly correlation compared to U-Net-AR: on KS at $H{=}400$, RMSE drops by $62.8\%$ and ACC improves by $85.1\%$, while the spectral-energy discrepancy decreases by $67.6\%$; on L96 at $H{=}100$, MSR-HINE achieves a $27.0\%$ RMSE reduction, a $46.8\%$ ACC improvement, and a $65.1\%$ reduction in spectral-energy error. These gains indicate that the multi-rate recurrent hierarchy better preserves slow-manifold structure while the latent fusion/correction steps mitigate compounding errors and distribution shift, yielding rollouts that remain both accurate and physically consistent further into the forecast window.

Despite these improvements, forecasting chaotic dynamics remains fundamentally limited by sensitivity to initial conditions; performance degrades at sufficiently long horizons as trajectories decorrelate. Promising directions include extending MSR-HINE to higher-dimensional PDEs, coupling the latent hierarchy with data assimilation to correct drift online, and learning adaptive temporal rates and gating schedules to further improve robustness across regimes.

\begin{table*}[t]
\centering
\small
\setlength{\tabcolsep}{4.5pt}
\renewcommand{\arraystretch}{1.15}
\resizebox{\textwidth}{!}{%
\begin{tabular}{ll l c c c c c c}
\toprule
Dataset & $H$ & Model &
RMSE@H (mean $\pm$ std) &
ACC@H (mean $\pm$ std) &
SpecErr (mean $\pm$ std) &
$\Delta$RMSE (\%) &
$\Delta$ACC (\%) &
$\Delta$SpecErr (\%) \\
\midrule
KS  & 400 & \textbf{MSR-HINE} & \textbf{1.094e+00 $\pm$ 3.525e-01} & \textbf{0.828 $\pm$ 0.109} & \textbf{5.384e-02 $\pm$ 2.889e-02} & \textbf{+62.8} & \textbf{+85.1} & \textbf{+67.6} \\
KS  & 400 & HINE-L2  & 1.920e+00 $\pm$ 4.109e-01 & 0.500 $\pm$ 0.209 & 1.435e-01 $\pm$ 3.469e-02 & +34.7 & +56.7 & +13.8 \\
KS  & 400 & U-Net-AR  & 2.941e+00 $\pm$ 1.770e-01 & -0.155 $\pm$ 0.124 & 1.663e-01 $\pm$ 6.031e-02 & +0.0 & +0.0 & +0.0 \\
\midrule
L96 & 100 & \textbf{MSR-HINE} & \textbf{3.445e+00 $\pm$ 7.997e-01} & \textbf{0.545 $\pm$ 0.193} & \textbf{1.065e-01 $\pm$ 3.685e-02} & \textbf{+27.0} & \textbf{+46.8} & \textbf{+65.1} \\
L96 & 100 & HINE-L2  & 4.982e+00 $\pm$ 5.849e-01 & 0.091 $\pm$ 0.172 & 3.379e-01 $\pm$ 7.517e-02 & -5.6 & -6.2 & -10.7 \\
L96 & 100 & U-Net-AR  & 4.718e+00 $\pm$ 4.632e-01 & 0.144 $\pm$ 0.161 & 3.051e-01 $\pm$ 6.480e-02 & +0.0 & +0.0 & +0.0 \\
\bottomrule
\end{tabular}%
}
\caption{\textbf{Long-horizon summary metrics and relative gains over U-Net-AR.}
We report RMSE and ACC at the terminal horizon ($H{=}400$ for KS, $H{=}100$ for L96) and a spectral/energy fidelity error (\emph{SpecErr}, lower is better; see Sec.~\ref{sec:eval_metrics}). Percentage changes $\Delta$ are computed relative to U-Net-AR (positive indicates improvement). Bold denotes the best value per dataset/metric.}
\label{tab:conclusion_summary}
\end{table*}

\section*{Acknowledgments}
This work was supported in part by the AFOSR Grant FA9550-24-1-0327.
\bibliographystyle{unsrt} 

\bibliography{manuscript}

\end{document}